\pdfoutput=1

\documentclass[11pt]{article}

\usepackage[final]{acl}

\usepackage{times}
\usepackage{latexsym}

\usepackage[T1]{fontenc}

\usepackage[utf8]{inputenc}

\usepackage{microtype}

\usepackage{inconsolata}

\usepackage{graphicx}
\usepackage[ruled,linesnumbered]{algorithm2e}
\usepackage{amsmath}
\usepackage{amsthm}
\usepackage{enumitem}
\setlist{left=\parindent} 
\usepackage{xcolor}
\definecolor{lightfreshblue}{RGB}{202, 224, 239}
\usepackage{graphicx}
\usepackage{booktabs}
\usepackage{multirow}
\usepackage{xspace}
\usepackage{balance}
\usepackage{bbold}
\usepackage{pifont}
\usepackage{colortbl}
\usepackage[normalem]{ulem}
\useunder{\uline}{\ul}{}
\usepackage{tikz}
\usepackage{array}
\usepackage{longtable}
\usepackage{subcaption}
\usepackage{float}
\usepackage{placeins}
\usepackage{pdfpages}
\usepackage[most]{tcolorbox}
\usepackage{lipsum}
\theoremstyle{definition}

\newcommand{\eg}{\emph{e.g.,}\xspace}

\newcommand{\stitle}[1]{\noindent{\bf #1}\hspace{1.5pt}}

\newcommand{\sys}{\textsc{ARise}\xspace}

\usepackage{etoolbox}
\makeatletter
\patchcmd{\@setref}{\bfseries ??}{\bfseries \color{red} ??}{}{}
\patchcmd{\NAT@citex}{\bfseries ?}{\bfseries \color{red} ?}{}{}
\patchcmd{\NAT@citexnum}{\bfseries ?}{\bfseries \color{red} ?}{}{}
\ifdefined\HyRef@autosetref
  \patchcmd{\HyRef@autosetref}{\bfseries ??}{{\bfseries \color{red} ??}}{}{}
\fi
\makeatother

\title{\sys: Towards Knowledge-Augmented Reasoning via Risk-Adaptive Search}

\author{
\begin{tabular}{@{}ccc@{}}
Yize Zhang\textsuperscript{1,2,3}\thanks{Equal contribution.} &
Tianshu Wang\textsuperscript{5,7,8}\footnotemark[1]\thanks{Work done during an internship at SenseTime.} &
Sirui Chen\textsuperscript{1,6} \\
Kun Wang\textsuperscript{4} &
Xingyu Zeng\textsuperscript{4} &
Hongyu Lin\textsuperscript{5} \\
Xianpei Han\textsuperscript{5} &
Le Sun\textsuperscript{5} &
Chaochao Lu\textsuperscript{1,2}\thanks{Corresponding author.}
\end{tabular}
\\[1.5em]
\textsuperscript{1}Shanghai AI Laboratory \quad
\textsuperscript{2}Shanghai Innovation Institute \quad
\textsuperscript{3}Shanghai Jiao Tong University \\
\textsuperscript{4}SenseTime \quad
\textsuperscript{5}Institute of Software, Chinese Academy of Sciences \quad
\textsuperscript{6}Tongji University \\
\textsuperscript{7}Hangzhou Institute for Advanced Study, \textsuperscript{8}University of Chinese Academy of Sciences \\
\texttt{ez220523@sjtu.edu.cn, tianshu2020@iscas.ac.cn, luchaochao@pjlab.org.cn}
}

\begin{document}
\maketitle

\begin{abstract}

Large language models (LLMs) have demonstrated impressive capabilities and are receiving increasing attention to enhance their reasoning through scaling test-time compute.
However, their application in \emph{open-ended, knowledge-intensive, complex reasoning} scenarios is still limited.
Reasoning-oriented methods struggle to generalize to open-ended scenarios due to implicit assumptions of complete world knowledge.
Meanwhile, knowledge-augmented reasoning (KAR) methods fail to address two core challenges: 1) error propagation, where errors in early steps cascade through the chain, and 2) verification bottleneck, where the explore–exploit trade-off arises in multi-branch decision processes.
To overcome these limitations, we introduce \sys, a novel framework that integrates risk assessment of intermediate reasoning states with dynamic retrieval-augmented generation (RAG) within a Monte Carlo tree search paradigm. This approach enables effective construction and optimization of reasoning plans across multiple maintained hypothesis branches.
Experimental results show that \sys significantly outperforms the state-of-the-art KAR methods by up to 23.10\%, and the latest RAG-equipped large reasoning models by up to 25.37\%.
Our project page is at \href{https://opencausalab.github.io/ARise}{https://opencausalab.github.io/ARise}.

%

\end{abstract}
\section{Introduction}

Large language models (LLMs) have demonstrated impressive capabilities across a wide range of tasks~\cite{DBLP:journals/corr/abs-2303-08774,DBLP:journals/corr/abs-2303-18223,DBLP:journals/corr/abs-2303-12712}.
Despite their great success, LLMs still face fundamental challenges in complex reasoning scenarios, hindering their reliable application in real-world domains such as science, finance, and healthcare~\cite{DBLP:journals/corr/abs-2211-09085,DBLP:conf/icaif/LiWDC23,thirunavukarasu2023large}.
To address this gap, recent research has increasingly focused on enhancing LLM reasoning by scaling test-time compute to emulate System 2 slow thinking, moving beyond System 1 fast responses~\cite{kahneman2011thinking,DBLP:journals/corr/abs-2408-03314}.
Extensive efforts have led to various approaches, including prompt-based~\cite{DBLP:journals/csur/YuZTW24}, search-based~\cite{DBLP:conf/emnlp/HaoGMHWWH23}, and learning-based~\cite{OpenAI-o1,deepseek-r1}, showing great promise.

\begin{figure}[t]
    \centering
    \includegraphics[width=1\linewidth]{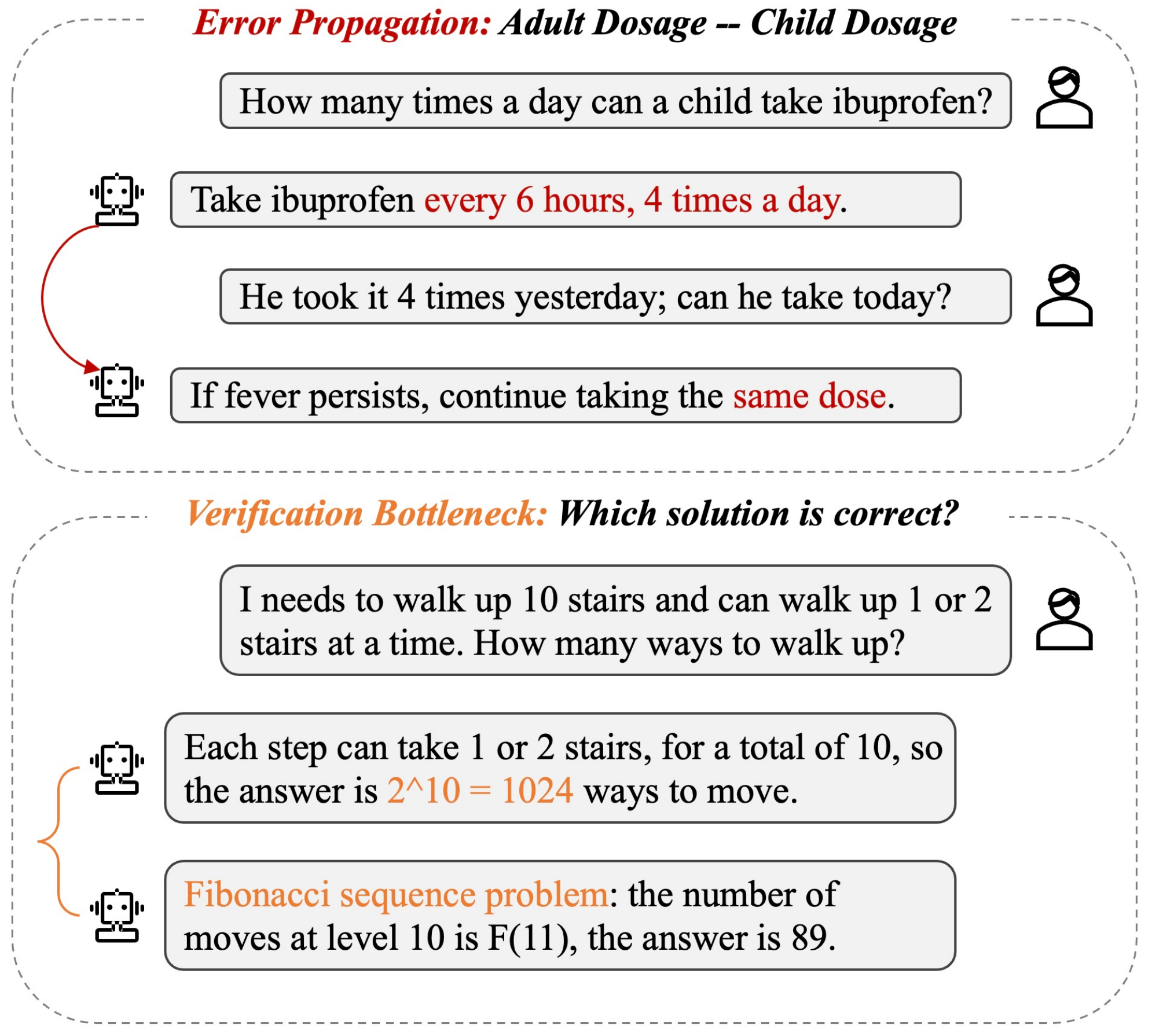}
    \caption{\textbf{Error propagation and verification bottleneck.} Prior knowledge-augmented reasoning works fail to address two core challenges: 1) error propagation, where errors in early steps cascade through the chain, and 2) verification bottleneck, where the explore–exploit trade-off arises in multi-branch decision processes.}
    \label{fig:1}
\end{figure}

However, reasoning-oriented methods struggle to generalize to open-ended scenarios~\cite{DBLP:journals/corr/abs-2206-10498,DBLP:conf/cikm/AmirizanianiMSM24}, primarily due to their implicit assumptions of complete world knowledge.
While these solutions like large reasoning models (LRMs) have achieved expert or superhuman performance on tasks such as math and code, their success relies heavily on clear standards for search algorithms or reinforcement learning~\cite{DBLP:conf/nips/ZhangZHYD024,xu2025largereasoningmodelssurvey}.
Such an exclusive focus on enhancing LLM reasoning implicitly assumes that LLMs already possess all the knowledge necessary for reasoning, which is often lacking in open-ended or domain-specific contexts.
For example, legal defense requires specialized jurisprudence knowledge, or medical diagnosis demands up-to-date clinical guidelines.
In fact, reasoning is a dynamic process of integrating multiple knowledge to draw conclusions~\cite{DBLP:journals/csur/YuZTW24,OpenAI-deep-research}, thus making knowledge acquisition an essential part of reasoning.

Meanwhile, current knowledge-augmented reasoning (KAR) methods, as illustrated in \autoref{fig:1}, are hindered by \emph{error propagation} and \emph{verification bottleneck}, which undermine reasoning reliability.
To acquire knowledge for reasoning, retrieval-augmented generation (RAG) has been shown to be an effective way of dynamically retrieving documents as intermediate results~\cite{DBLP:conf/nips/LewisPPPKGKLYR020,DBLP:journals/corr/abs-2410-02338}.
Prompt-based methods further extend KAR through chain-of-thought (CoT) prompting, which decomposes complex reasoning into sub-steps and iteratively retrieves relevant knowledge as reasoning proceeds~\cite{DBLP:conf/acl/ZhaoLJQB23,DBLP:journals/corr/abs-2411-19443,DBLP:conf/iclr/LiZCDJPB24}.
However, this approach is plagued by error propagation, where errors in early steps can cascade through the chain.
While search-based methods can mitigate error propagation by maintaining multiple hypothesis branches, verification bottleneck limits the effective explore–exploit trade-off in KAR.
Existing verification solutions remain unsatisfactory as they rely on error-prone self-verification~\cite{DBLP:journals/corr/abs-2402-08115,DBLP:conf/iclr/0002WSLCNCZ23,DBLP:journals/corr/abs-2406-02746}, or on specific verifier training~\cite{DBLP:journals/corr/abs-2410-08146,DBLP:conf/nips/ZhangZHYD024}. 

To overcome these limitations, we present a novel framework, \sys, towards knowledge-\textbf{\textsc{A}}ugmented \textbf{\textsc{R}}easoning via r\textbf{\textsc{i}}sk-adaptive \textbf{\textsc{se}}arch.
As shown in \autoref{fig:pipeline}, \sys consists of three components: \textbf{reasoning state generation}, \textbf{Monte Carlo tree search} (MCTS), and \textbf{risk assessment}.
Specifically, \sys iteratively refines reasoning steps through decomposition, retrieval-then-reasoning to provide fine-grained knowledge for LLMs (\autoref{sec:mcts-trajectory}). 
MCTS treats each step as a node in the search tree, expanding linear reasoning to mitigate error propagation by enabling focused exploration of promising reasoning states and allowing backtracking when necessary (\autoref{sec:mcts}).
Risk assessment leverages Bayesian risk minimization to evaluate the uncertainty of each state, dynamically balancing explore–exploit trade-off to guide the search towards both reliable and novel reasoning directions (\autoref{sec:rv}).
In this way, \sys enables robust and efficient complex reasoning by combining structured decomposition, knowledge retrieval, and risk-adaptive exploration in a unified framework.

We conducted comprehensive experiments with multiple LLMs on three challenging multi-hop question answering (QA) benchmarks that require complex reasoning and knowledge integration.
Experimental results demonstrate that \sys significantly outperforms the state-of-the-art (SOTA) KAR methods, with an average of 23.10\% and 15.52\% improvement in accuracy and F1.
In addition, when compared to the latest LRMs~\cite{deepseek-r1} equipped with RAG, \sys also improve the average accuracy and F1 of 4.04\% and 25.37\%.
These results verify the effectiveness of \sys for open-ended, knowledge-intensive, complex reasoning tasks.

To summarize, our \textbf{contributions} are as follows:
\begin{itemize}[nosep]
  \item We propose a knowledge-augmented framework for open-ended complex reasoning and design a risk-adaptive MCTS algorithm to balance explore-exploit trade-off for reasoning.
  \item We conduct comprehensive experiments to verify the effectiveness of \sys and to demonstrate that it outperforms the SOTA KAR methods and the latest LRMs equipped with RAG.
  \item We provide empirical insights that 1) search-based wide reasoning can explore more solutions than learning-based deep reasoning, and 2) \sys progressively approaches optimal performance through model size scaling.
\end{itemize}

\section{The \sys Method}
\begin{figure*}[h]
    \centering
    \includegraphics[width=1\linewidth]{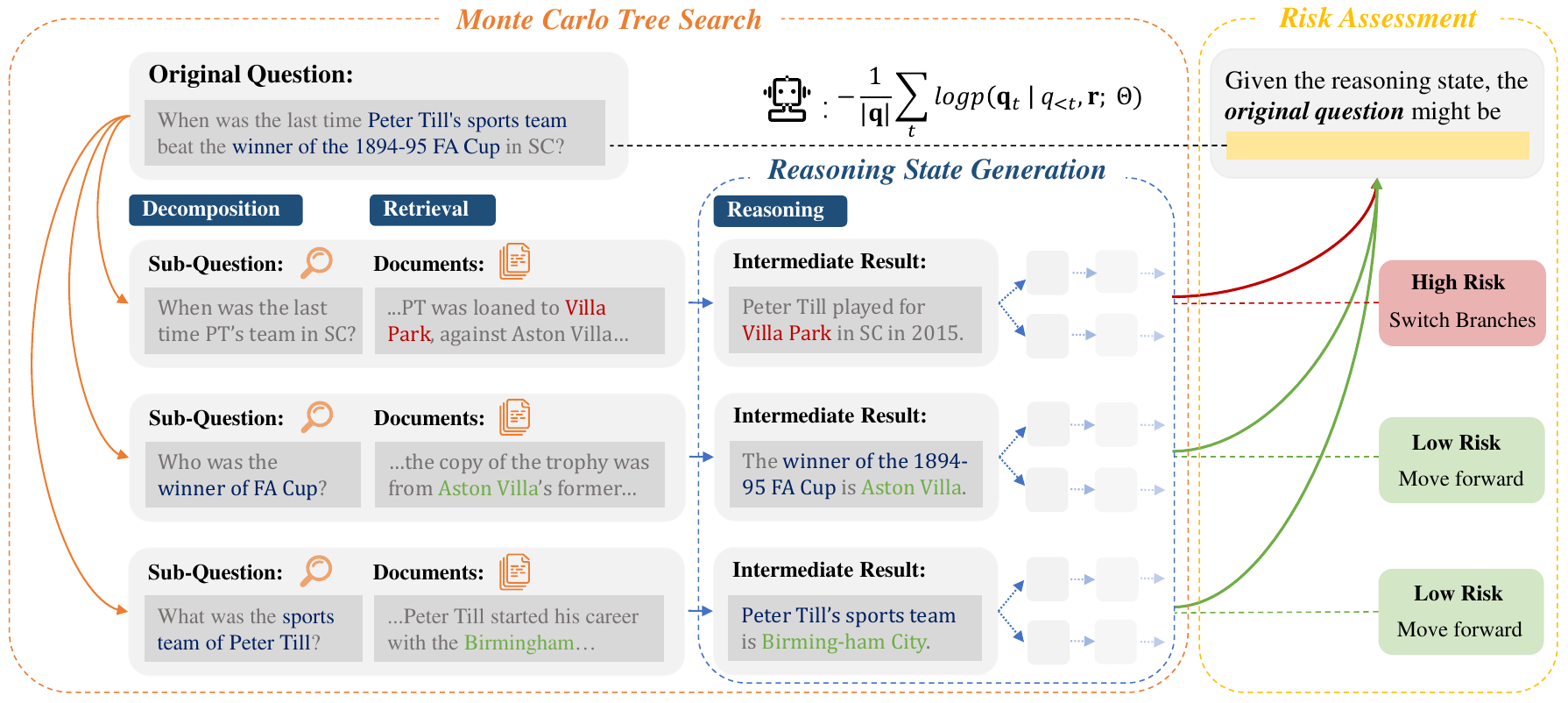}
    \caption{\textbf{Pipeline of \sys.} \sys iteratively refines reasoning steps through decomposition, retrieval-then-reasoning, providing fine-grained knowledge for LLMs (\autoref{sec:mcts-trajectory}). 
    MCTS treats each step as a node in the search tree, expanding linear reasoning to mitigate error propagation by enabling exploration of reasoning paths and allowing backtracking when necessary (\autoref{sec:mcts}).
    Risk assessment leverages Bayesian risk minimization to evaluate the quality of each state, dynamically optimizing action strategies to guide the search towards promising directions (\autoref{sec:rv}).}
    \vspace{-4mm}
    \label{fig:pipeline}
\end{figure*}

Our method, \sys, utilizes risk-adaptive tree search to provide the model with more external knowledge, thereby effectively enhancing its reasoning capabilities.
Our pipeline is illustrated in \autoref{fig:pipeline} and comprises the following three parts:

\begin{itemize}[wide,nosep]

\item{\textbf{Reasoning State Generation: }}
The single step of the policy model\footnote{We use ``policy models'' to refer to the LLMs employed during the inference phase.} consists of an action pair: decomposition and retrieval-then-reasoning.
Each step serves as a node, encoding an intermediate reasoning state.

\item{\textbf{Monte Carlo Tree Search: }}
MCTS transforms a sequence of interconnected nodes into a tree structure.
Each node can undergo a simulated rollout under guidance.
The local value incorporating future reward can be updated without incurring the cost of taking an actual forward step.
\item{\textbf{Risk Assessment: }}
We designed the Risk-Value function to assess the risks of the intermediate reasoning states at each node.
The policy model is capable of dynamically formulating and adjusting action strategies based on the actual risk associated with each branch.
\end{itemize}


\subsection{Reasoning State Generation}
\label{sec:mcts-trajectory}

To define steps more clearly and granularly, we prompt LLMs to perform problem decomposition and retrieval-then-reasoning in an interleaved manner.
These two consecutive actions together constitute a single step.
Intermediate results at each step are continuously appended to the entire reasoning state and serve as new inputs for subsequent steps, progressively approaching the final solution to complex tasks.
This approach, where intermediate subtasks and their labels are concatenated to the original task's input to form a new input sequence, is widely applied in compounded tasks~\citep{DBLP:conf/iclr/WiesLS23, DBLP:conf/acl/RajaniMXS19, DBLP:journals/corr/abs-2110-14168}.
Specifically, at the $i^{\text{th}}$ step\footnote{We use the \textbf{bold} notation for vectors and non-bold notation for scalars.}, the input comprises the original problem $\mathbf{q}$ and the intermediate results ${\mathbf{r_1}, \mathbf{r_2}, \mathbf{\dots}, \mathbf{r_{i-1}}}$ from previous steps, with the latter forming the reasoning state $\mathbf{s_{i-1}} = \mathbf{r_1} \oplus \mathbf{r_2} \oplus \mathbf{\dots} \oplus \mathbf{r_{i-1}}$.
The policy model then decomposes the problem into a subproblem $\mathbf{d_i}$, following the policy $\pi(\mathbf{d_i} \mid \mathbf{q}, \mathbf{s_{i-1}})$.
Based on the subproblem $\mathbf{d_i}$ and the retrieved documents, the intermediate result $\mathbf{r_i}$ is then generated and appended to the reasoning state repeatedly.
Each step encodes an $(\mathbf{s_{i-1}}, \mathbf{a_i})$ pair, where $\mathbf{s_{i-1}}$ represents the state, and $\mathbf{a_i}$ is a set $\{\mathbf{d_i}, \mathbf{r_i}\}$ that implicitly reflects the step's two actions, with $\mathbf{d_i}$ being the outcome corresponding to the decomposition and $\mathbf{r_i}$ to the retrieval-then-reasoning.
A sequence of coherent steps, extending until the endpoint, collectively forms a complete trajectory.

\subsection{Monte Carlo Tree Search}
\label{sec:mcts}

The MCTS algorithm expands a single trajectory into a search tree structure.
The whole process begins with the original problem as the root node, followed by iterative searches consisting of selection, expansion, simulation, and backpropagation.
The four phases are detailed as follows:

\stitle{Selection.}
Starting from the root node and traversing the existing tree structure, the algorithm selects the optimal child node in preparation for the next expansion phase. To balance exploration and exploitation, the well-known Upper Confidence Bounds (UCT)~\citep{DBLP:conf/ecml/KocsisS06} is used in the selection process, formulated as:
\[\text{UCT}(\mathbf{s}, \mathbf{a}) = Q(\mathbf{s}, \mathbf{a}) + w\sqrt{\frac{\ln N(Pa(\mathbf{s}))}{N(\mathbf{s}, \mathbf{a})}},\]
where $N(\mathbf{s},\mathbf{a})$ and $N(Pa(\mathbf{s}))$ represent the visit counts of the current node and its parent node in previous searches, respectively.
The initial value of $Q(\mathbf{s}, \mathbf{a})$ is calculated by the Risk-Value function (detailed in \autoref{sec:rv}) and is subsequently updated during the backpropagation phase.

\stitle{Expansion.}
The model decomposes the original problem based on the reasoning state from different perspectives to generate new subproblems.
Each subproblem and its corresponding result form a distinct child node, which is then appended to the selected node, thereby expanding the tree in both width and depth.

\stitle{Simulation.}
The model initiates an imagined rollout from the selected node, proceeding until it reaches a leaf node.
This phase assists in assigning the current node a more farsighted value that incorporates future rewards by completing the imagined reasoning trajectory without altering the tree structure.
Within a single rollout, the model can still sample multiple times and greedily advance towards the leaf nodes.

\stitle{Backpropagation.}
The backpropagation phase updates the values of all nodes along the selected reasoning branch.
This process follows a bottom-up manner, where the parent node's value is determined by the values and visit counts of its child nodes.
The mathematical formulation is as follows:
\[Q(\mathbf{s}, \mathbf{a}) = \frac{\sum_{\mathbf{c} \in \mathcal{C}(\mathbf{s}, \mathbf{a})} Q(\mathbf{c}) \cdot N(\mathbf{c})}{\sum_{\mathbf{c} \in \mathcal{C}(\mathbf{s}, \mathbf{a})} N(\mathbf{c})},\]
where $\mathcal{C}(\mathbf{s}, \mathbf{a})$ denotes all child nodes of $(\mathbf{s}, \mathbf{a})$.

After reaching the predetermined number of search iterations, the tree structure and node values stabilize.
Ultimately, the model selects the optimal path by maximizing value at each step, following the greedy policy.

\subsection{Risk Assessment}
\label{sec:rv}

In this section, we delve into the Risk-Value function, which assesses the risks of reasoning states to guide the tree-search process.
To begin with, for a composite problem $\mathbf{q}$, we treat its decomposition and retrieval-then-reasoning as a statistical decision of a probabilistic process~\citep{DBLP:journals/ipm/ZhaiL06, DBLP:conf/sigir/LaffertyZ01}.
Specifically, given a set of decomposed subproblems $D = \{\mathbf{d^1}, \mathbf{d^2}, \mathbf{\dots}, \mathbf{d^k}\}$ and the corresponding set of intermediate results $R = \{\mathbf{r^1}, \mathbf{r^2}, \mathbf{\dots}, \mathbf{r^k}\}$\footnote{In this notation, the subscript of the symbol denotes the sequence number of the reasoning step, while the superscript indicates the identifier for different reasoning perspectives.}, the quality of a node state can be evaluated using a relevance score $p(\mathbf{r} \mid \mathbf{q}), \mathbf{r} \in R$~\citep{DBLP:conf/emnlp/SachanLJAYPZ22}.
We substitute the ``problem generation likelihood''~\citep{DBLP:conf/sigir/ZhaiL01, DBLP:conf/sigir/PonteC98} as an alternative to the relevance score after applying the Bayes' rule:
\[\log p (\mathbf{r} \mid \mathbf{q}) = \log p(\mathbf{q} \mid \mathbf{r}) + \log p(\mathbf{r}) + c,\]
where $p(\mathbf{r})$ is the prior belief that $\mathbf{r}$ is relevant to any problem and is assumed to be uniform in this case.
We can also drop $c$ since it is the intermediate result-independent constant.
The formula is then reduced to:
\[\log p (\mathbf{r} \mid \mathbf{q}) \propto \log p(\mathbf{q} \mid \mathbf{r}), \forall \mathbf{r} \in R,\]
where $p(\mathbf{q} \mid \mathbf{r})$ captures how well the intermediate results $\mathbf{r}$ fit the particular problem $\mathbf{q}$.
We utilize the policy model to compute the average log-likelihood of generating the original problem tokens in order to estimate $\log p(\mathbf{q} \mid \mathbf{r})$~\citep{DBLP:conf/emnlp/SachanLJAYPZ22, DBLP:conf/emnlp/YuanYWZL24}, and define the expected risk of a node $(\mathbf{s},\mathbf{a})$ pointing to $\mathbf{r}$ as follows:
\[
\resizebox{\columnwidth}{!}{$\text{Risk}((\mathbf{s},\mathbf{a})\rightarrow \mathbf{r} \mid \mathbf{q})=-\frac{1}{|\mathbf{q}|}\sum\limits_t \log p(q_t \mid \mathbf{q}_{<t}, \mathbf{r}; \Theta),$}
\]
where $|\mathbf{q}|$ denotes the length of the original problem, $q_t$ represents the $t^{th}$ token in $\mathbf{q}$, and $\mathbf{q}_{<t}$ refers to the sequence of tokens preceding the $t^{th}$ token in $\mathbf{q}$. 
$\Theta$ denotes the parameters of the policy model.
Finally, the risk is scaled to the range (0, 1) through a sigmoid function in the opposite direction, serving as the node value:
\[Q(\mathbf{s},\mathbf{a})=1-\frac{1}{1+e^{\alpha\cdot(\text{Risk}-\beta)}},\]
where $\alpha, \beta$ are the translation and scaling factors.

\section{Experiments}
\label{sec:exp-trajectory}

\begin{table*}[t]
\centering
\small
\renewcommand{\arraystretch}{1.2} 
\setlength{\tabcolsep}{3pt} 
\begin{tabular}{c >{\centering\arraybackslash}p{1.2cm} >{\centering\arraybackslash}p{1.2cm}
                                >{\centering\arraybackslash}p{1.2cm} >{\centering\arraybackslash}p{1.2cm}
                                >{\centering\arraybackslash}p{1.2cm} >{\centering\arraybackslash}p{1.2cm}
                                >{\centering\arraybackslash}p{1.2cm} >{\centering\arraybackslash}p{1.2cm}}
\toprule
\multirow{2}{*}{\textbf{Method}} & \multicolumn{2}{c}{\textbf{HotpotQA}} & \multicolumn{2}{c}{\textbf{2Wiki}} & \multicolumn{2}{c}{\textbf{MusiQue}} & \multicolumn{2}{c}{\textbf{Average}} \\ 
\cmidrule(lr){2-3} \cmidrule(lr){4-5} \cmidrule(lr){6-7} \cmidrule(lr){8-9}  
                        & \textbf{EM}  & \textbf{F1}  & \textbf{EM}  & \textbf{F1}  & \textbf{EM}  & \textbf{F1}  & \textbf{EM}  & \textbf{F1}  \\ 
\midrule
\rowcolor[HTML]{E0F0FF} 
\multicolumn{9}{c}{\textbf{\textit{Qwen2.5-14B-Instruct}}} \\
Vanilla                 & 59.50  & 63.63  & 37.00  & 50.33  & 14.50  & 47.07  & 37.00  & 53.68  \\  
\rowcolor[HTML]{F0F0F0} 
\multicolumn{9}{c}{\textit{Prompt-based}} \\
Query2Doc~\citep{DBLP:conf/emnlp/WangYW23}              & 61.00  & 67.65  & 38.00  & 53.40  & 22.00  & 55.79  & 40.33  & 58.95  \\  
Self-Ask~\citep{DBLP:conf/emnlp/PressZMSSL23}           & 58.50  & 64.74  & 38.50  & 53.45  & 25.00  & 58.59  & 40.67  & 58.93  \\  
Verify-and-Edit~\citep{DBLP:conf/acl/ZhaoLJQB23}        & 62.50  & 70.16  & 38.50  & 57.68  & 22.00  & 55.30  & 41.00  & 61.05  \\  
Auto-RAG~\citep{DBLP:journals/corr/abs-2411-19443}      & 68.00  & 66.64  & 53.00  & 55.13  & 35.50  & 59.05  & 52.17  & 60.27  \\  
\rowcolor[HTML]{F0F0F0} 
\multicolumn{9}{c}{\textit{Search-based}} \\
CoT-SC~\citep{DBLP:conf/iclr/0002WSLCNCZ23}            & 63.00  & 71.39  & 37.50  & 54.41  & 16.00  & 57.46  & 38.83  & 61.09  \\  
RATT~\citep{DBLP:journals/corr/abs-2406-02746}         & 64.50  & 73.91  & 43.00  & 57.48  & 24.00  & 63.76  & 43.83  & 65.05  \\
\rowcolor[HTML]{E4F8E4} 
\textbf{\sys (ours)}    & \textbf{73.50}  & \textbf{75.39}  & \textbf{56.50}  & \textbf{62.61}  & \textbf{40.50}  & \textbf{65.87}  & \textbf{56.83}  & \textbf{67.96}  \\
\midrule
\rowcolor[HTML]{E0F0FF} 
\multicolumn{9}{c}{\textbf{\textit{Qwen2.5-7B-Instruct}}} \\
Vanilla                 & 54.50  & 63.63  & 36.50  & 50.33  & 11.00  & 47.07  & 34.00  & 53.68  \\  
\rowcolor[HTML]{F0F0F0} 
\multicolumn{9}{c}{\textit{Prompt-based}} \\
Query2Doc~\citep{DBLP:conf/emnlp/WangYW23}              & 60.00  & 67.31  & 37.00  & 53.78  & 14.50  & 54.59  & 37.17  & 58.56  \\  
Self-Ask~\citep{DBLP:conf/emnlp/PressZMSSL23}           & 44.00  & 61.43  & 27.50  & 48.86  & 22.00  & 57.48  & 31.17  & 55.92  \\  
Verify-and-Edit~\citep{DBLP:conf/acl/ZhaoLJQB23}        & \textbf{67.00}  & 69.87  & 39.00  & 53.91  & 21.50  & 54.75  & 42.50  & 59.51  \\  
Auto-RAG~\citep{DBLP:journals/corr/abs-2411-19443}      & 66.50  & 66.33  & 44.50  & 54.00  & \textbf{29.50}  & 57.19  & 46.83  & 59.17  \\  
\rowcolor[HTML]{F0F0F0} 
\multicolumn{9}{c}{\textit{Search-based}} \\
CoT-SC~\citep{DBLP:conf/iclr/0002WSLCNCZ23}            & 60.50  & 70.66  & 38.00  & 54.01  & 15.00  & 56.72  & 37.83  & 60.46  \\  
RATT~\citep{DBLP:journals/corr/abs-2406-02746}         & 58.50  & 68.88  & 36.50  & 53.91  & 18.00  & 56.58  & 37.67  & 59.79  \\
\rowcolor[HTML]{E4F8E4} 
\textbf{\sys (ours)}    & 66.50  & \textbf{73.87}  & \textbf{47.50}  & \textbf{61.37}  & 29.00  & \textbf{62.26}  & \textbf{47.67}  & \textbf{65.83}  \\ 
\midrule
\rowcolor[HTML]{E0F0FF} 
\multicolumn{9}{c}{\textbf{\textit{Llama3.1-8B-Instruct}}} \\
Vanilla                 & 55.50  & 63.63  & 31.50  & 50.33  & 14.00  & 47.07  & 33.67  & 53.68  \\  
\rowcolor[HTML]{F0F0F0} 
\multicolumn{9}{c}{\textit{Prompt-based}} \\
Query2Doc~\citep{DBLP:conf/emnlp/WangYW23}              & 57.50  & 63.52  & 32.50  & 49.78  & 19.00  & 49.91  & 36.33  & 54.40  \\  
Self-Ask~\citep{DBLP:conf/emnlp/PressZMSSL23}           & 57.00  & 62.10  & 40.00  & 51.26  & 20.50  & 52.14  & 39.17  & 55.17  \\  
Verify-and-Edit~\citep{DBLP:conf/acl/ZhaoLJQB23}        & 50.50  & 65.07  & 29.00  & 51.91  & 13.50  & 49.77  & 31.00  & 55.58  \\  
Auto-RAG~\citep{DBLP:journals/corr/abs-2411-19443}      & 51.00  & 53.37  & 35.00  & 48.81  & 21.50  & 52.61  & 35.83  & 51.60  \\  
\rowcolor[HTML]{F0F0F0} 
\multicolumn{9}{c}{\textit{Search-based}} \\
CoT-SC~\citep{DBLP:conf/iclr/0002WSLCNCZ23}            & \textbf{64.50}  & 71.40  & 45.00  & 58.02  & 22.00  & 59.43  & 43.83  & 62.96  \\  
RATT~\citep{DBLP:journals/corr/abs-2406-02746}         & 58.50  & 71.18  & \textbf{46.00}  & 56.18  & \textbf{29.50}  & 63.66  & \textbf{44.67}  & 63.67  \\
\rowcolor[HTML]{E4F8E4} 
\textbf{\sys (ours)}    & 63.00  & \textbf{74.78}  & 34.50  & \textbf{63.19}  & 24.50  & \textbf{66.38}  & 40.67  & \textbf{68.12}  \\ 
\bottomrule
\end{tabular}
\caption{\textbf{Comparison of \sys with a wide range of baselines.}}
\label{tab:main}
\end{table*}


\subsection{Setup}

\paragraph{Datasets.}
We use HotpotQA~\citep{DBLP:conf/emnlp/Yang0ZBCSM18}, 2WikiMultihopQA~\citep{DBLP:conf/coling/HoNSA20}, and MusiQue~\citep{DBLP:journals/tacl/TrivediBKS22} as test sets.
These datasets span a wide range of topics, necessitating the retrieval and reasoning over multiple supporting documents.
To balance computational efficiency and evaluation robustness, we conduct experiments on a subset of 200 randomly selected questions \citep{DBLP:journals/corr/abs-2412-12881, feng2025airrag}.
More details are available in \autoref{sec:appd}.

\paragraph{Baselines and Metrics.}

All baselines are incorporated with RAG.
For prompt-based baselines, we compare \sys with Query2Doc~\citep{DBLP:conf/emnlp/WangYW23}, Self-Ask~\citep{DBLP:conf/emnlp/PressZMSSL23}, Verify-and-Edit~\citep{DBLP:conf/acl/ZhaoLJQB23} and Auto-RAG~\citep{DBLP:journals/corr/abs-2411-19443}.
For search-based baselines, we use Self-Consistency (SC)~\citep{DBLP:conf/iclr/0002WSLCNCZ23} and Retrieval-augmented-thought-tree (RATT)~\citep{DBLP:journals/corr/abs-2406-02746}.
We choose EM accuracy and F1 score as our evaluation metrics.
The prediction is correct if the ground truth answer is exactly contained \citep{DBLP:journals/corr/abs-2412-12881, feng2025airrag}.
More details are available in \autoref{sec:appm}.

\paragraph{Implementation Details.}
In the main experiments, we configure the number of iterations to 200, the exploration weight factor $w$ in UCT to 1.4, and the temperature to 0.7.
For the retrieval process, we employ BM25 as our retriever.
Prompts for forward reasoning directly describe the task with zero-shot instructions.
For verification prompts, we provide few-shot demonstrations.
Further details and specific prompts are available in \autoref{sec:appp}.

\subsection{Main Results}

\paragraph{\textit{Finding 1: }\sys demonstrates superior performance.}
\autoref{tab:main} presents the comprehensive experimental results on \sys and various baselines.
Specifically, on the Qwen2.5-14B-Instruct model, \sys outperforms across all benchmarks, achieving an absolute improvement of 19.83\% in EM over the vanilla RAG method, 13.29\% over prompt-based baselines, and 15.5\% over search-based baselines.
\sys maintains robust performance on the Qwen2.5-7B-Instruct model with an absolute improvement of 13.67\% in EM over the vanilla RAG method and overall surpasses various baselines.
We observed that \sys performs slightly worse on Llama models.
To further analyze this, we identify another interesting phenomenon: Auto-RAG, which adopts the same interleaved decomposition and retrieval paradigm as \sys, also exhibits a decline on Llama.
This phenomenon suggests that the Llama model may not be well-suited for iterative problem decompositions.
In contrast, continuous-step reasoning methods such as CoT and tree-of-thoughts show better results.
Nevertheless, \sys still maintains a notable F1 advantage on Llama, indicating its effectiveness in selecting more promising paths.
%


\paragraph{\textit{Finding 2: }\sys demonstrates substantial potential on more challenging tasks.}
We are excited to show that \sys demonstrates superior performance on more challenging datasets.
Based on average performance, the difficulty level increases progressively from HotpotQA to 2WikiMultihopQA, and then to MusiQue.
In particular, on the 14B model, \sys achieves a relative improvement of 23.53\% in EM over the vanilla RAG method on HotpotQA, which surges to 52.70\% and 179.31\% on 2WikiMultihopQA and MusiQue, respectively.
In contrast, a wide range of baselines show only average performance improvements of 5.74\%, 11.94\%, and 66.09\%, respectively.
The F1 score reflects the same trend, with a relative improvement of 18.48\%, 24.40\%, and 39.94\%, corresponding to the three benchmarks.

\subsection{Comparison to Learning-based LRMs}
\begin{figure}
    \centering
    \includegraphics[width=1\linewidth]{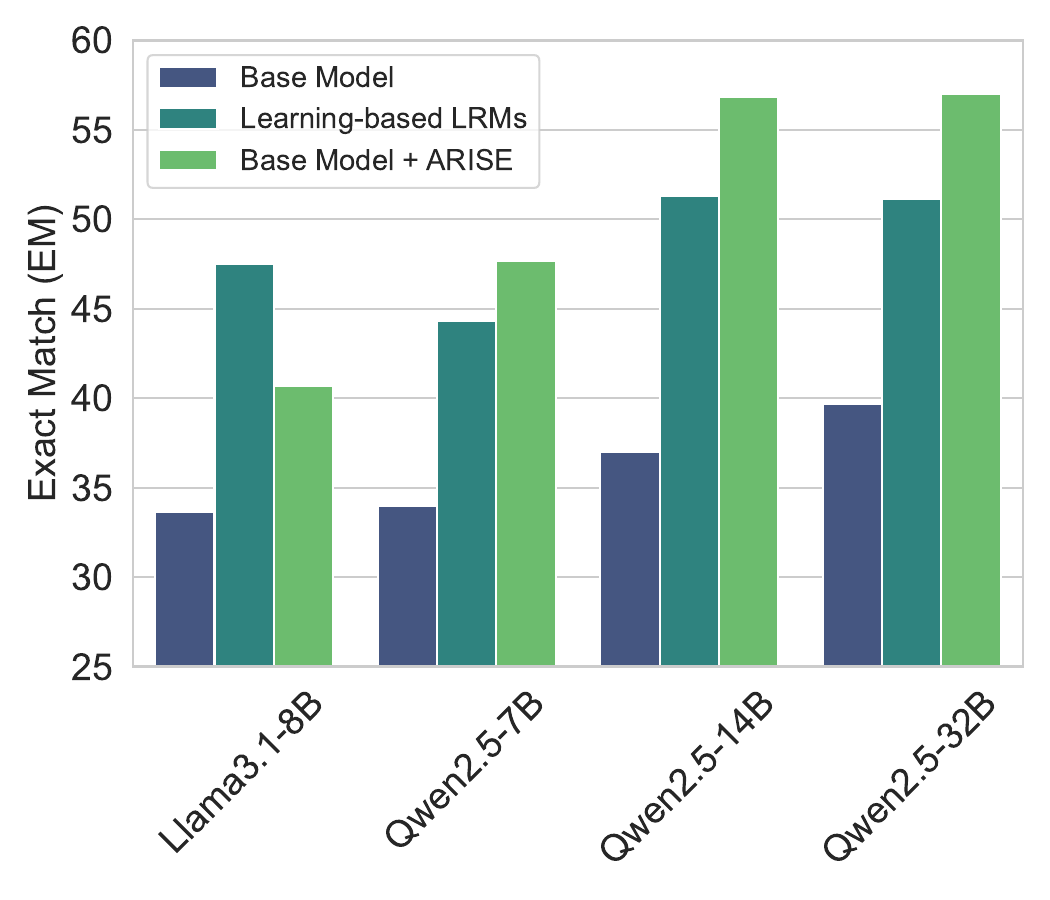}
    \caption{
    \textbf{Search-based reasoning vs. learning-based LRMs.} Learning-based LRMs like DeepSeek-R1 distilled models have not yet approached the point where they can effectively match or even replace search-based reasoning methods in terms of performance.
    }
    \label{fig:deepseek}
\end{figure}

\paragraph{\textit{Finding 3: }Learning-based LRMs have not yet approached the point where they can effectively match or even replace search-based reasoning as \sys.}
Our empirical comparison between base models with \sys and the DeepSeek-R1 distilled models reveals key insights into the effectiveness of test-time search.
These learning-based LRMs extract similar reasoning patterns from DeepSeek-R1.
As shown in \autoref{fig:deepseek}, \sys exhibits a performance advantage over the LRMs, especially on the Qwen model series.
While \sys slightly underperforms in comparison to DeepSeek-R1-style reasoning pattern on Llama, it consistently outperforms it on the Qwen models.
On average, \sys shows a relative improvement of 4.03\%, emphasizing the benefit of our search-based method.
These results suggest that while learning-based LRMs like DeepSeek-R1 distilled models provide valuable insights, they have not yet demonstrated the same effectiveness as search-based reasoning.

\subsection{Model Scale}
\label{sec:scale}
\begin{figure}[h]
    \centering
    \includegraphics[width=1\linewidth]{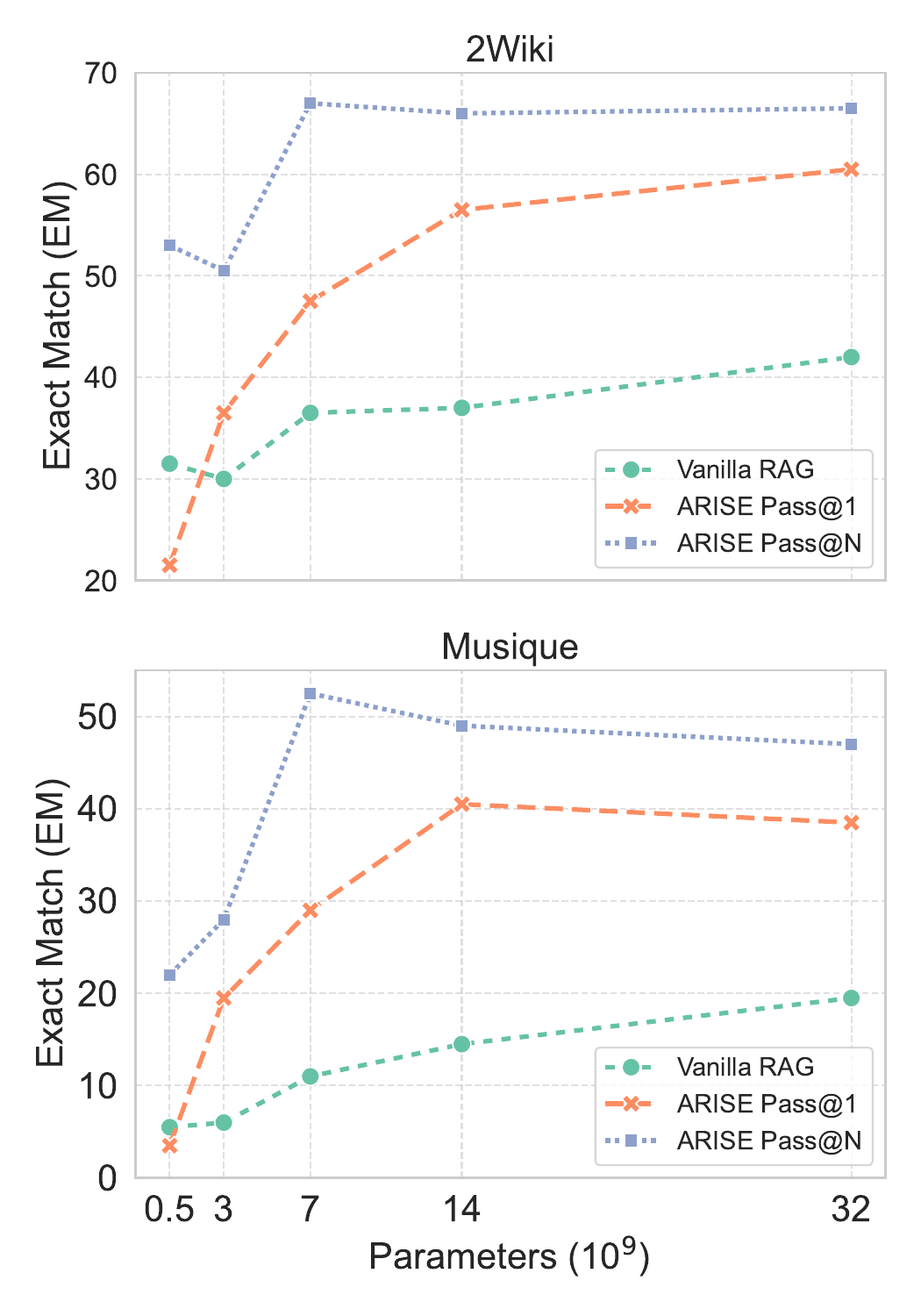}
    \caption{
    \textbf{Performance vs. model scales.} Although scaling up the model size shows diminishing returns for vanilla approaches, \sys better harnesses the potential of larger models in solving complex tasks.
    }
    \label{fig:scale}
\end{figure}

\paragraph{\textit{Finding 4: }\sys progressively approaches the optimal performance upper bound as model scale increases, unlocking the potential of larger models.} 

We conducted experiments on the Qwen2.5 series models, spanning a range of parameter scales from 0.5B to 32B, as illustrated in \autoref{fig:scale}.
To realistically and prominently reflect the related trends, we selected two more challenging test sets: 2WikiMultihopQA and Musique.
We employ Pass@N as the metric to evaluate the upper bound of the success rate, where a problem is considered solved as long as a single surviving node in the tree leads to the correct answer.
Pass@1, on the other hand, represents the success rate of the optimal path selected under the guidance of \sys.
The results show that Pass@N and the vanilla method exhibit similar trends as model parameters scale, but there is an average of 26.85\% significant room for improvement.
This indicates that appropriately scaling up inference computation offers substantial potential for enhancing performance.
We observed that after the model size reaches 7B parameters, the performance of both the upper bound and naive retrieval tends to saturate, suggesting diminishing returns with further scaling of model parameters.
In contrast, \sys demonstrates consistent improvement as model scale increases.
The accuracy of the optimal path selection gradually approaches the upper bound, with the success rate gap between Pass@1 and Pass@N decreasing from 25.00\% to 7.25\%.
%

\subsection{Computational Overhead}
The computational overhead is measured in terms of reasoning time (in minutes).

\subsubsection{Overhead in Relation to Search Space.}

\paragraph{\textit{Finding 5: }Moderate search space expansion boosts performance over minimal settings, but returns diminish rapidly as overhead surge.}

\begin{figure}[h]
    \centering
    \includegraphics[width=1\linewidth]{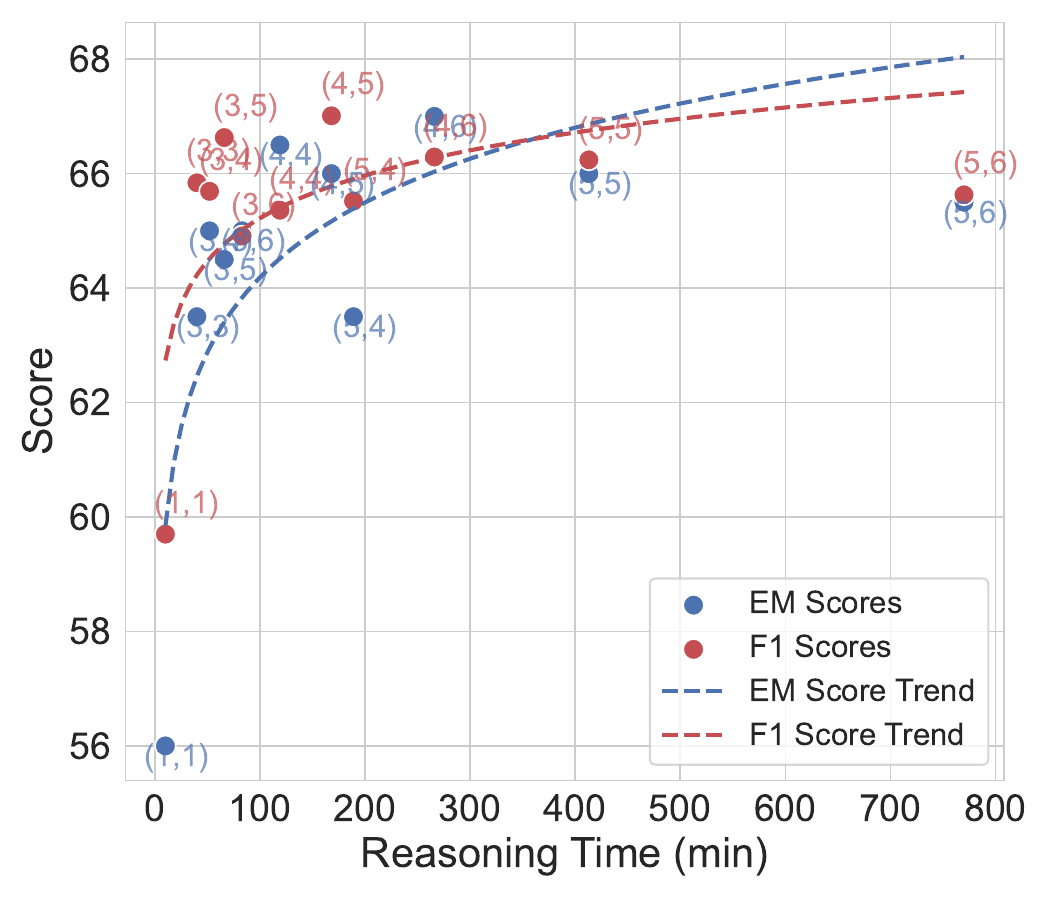}
    \caption{
    \textbf{Computational overhead in relation to the search space.} As the search space expands in depth and width, performance improves but with diminishing returns and significantly increased reasoning time. The (4,5) setting achieves a good trade-off and is adopted in our main experiments.
    }
    \label{fig:cost}
\end{figure}
We conducted experiments on the HotpotQA dataset using the Llama3.1-8B-Instruct model.
The results are under different search space configurations—defined by search depth and width.
As shown in \autoref{fig:cost}, expanding the search space leads to performance improvements (in EM and F1 scores), especially when increasing from a vanilla setting (depth 1, width 1) to moderate configurations. 
However, these improvements diminish as the depth and width increase further, while the computational overhead escalates rapidly. 
For instance, increasing the depth from 3 to 4 and the width from 4 to 6 results in a reasoning time increase from 52 to 266 minutes, with only marginal performance gain.
Based on this observation, we adopted a configuration of depth = 4 and width = 5 for our main experiments, which strikes a practical balance between accuracy and computational cost.
It is worth noting that the optimal search space configuration may vary across instances. 
Different questions demand varying depths of reasoning and degrees of knowledge integration, often corresponding to different numbers of reasoning hops. 
Thus, adaptively determining the search space based on task complexity remains an open research problem. 
Even state-of-the-art reasoning models still face the challenge of choosing between wider and deeper reasoning paths. 
We discuss this further in \autoref{sec:lim} and identify it as a promising direction for future work.

\subsubsection{Overhead Comparison with Baselines.}
\begin{table}[h]
\centering
\small
\renewcommand{\arraystretch}{1.2}
\setlength{\tabcolsep}{3pt}
\begin{tabular}{>{\centering\arraybackslash}p{2.2cm}
                >{\centering\arraybackslash}p{1.2cm}
                >{\centering\arraybackslash}p{1.2cm}
                >{\centering\arraybackslash}p{1.8cm}}
\toprule
\textbf{Method} & \textbf{EM} & \textbf{F1} & \textbf{Time (min)} \\
\midrule
Vanilla             & 14.50 & 47.07 & 10 \\
Query2Doc           & 22.00 & 55.79 & 16 \\
Self-Ask            & 25.00 & 58.59 & 18 \\
Verify-and-Edit     & 22.00 & 55.30 & 21 \\
Auto-RAG            & 35.50 & 59.05 & 26 \\
CoT-SC              & 16.00 & 57.46 & 69 \\
RATT                & 24.00 & 63.76 & 155 \\
\textbf{\sys (ours)} & \textbf{40.50} & \textbf{65.87} & \textbf{160} \\
\bottomrule
\end{tabular}
\caption{\textbf{Computational overhead comparison with baselines.} \sys achieves the best performance in both EM and F1 scores but incurs higher reasoning time due to its search-based multi-step reasoning. Practical deployment should consider the trade-off between computational resources and desired performance.}
\label{tab:baseline_comparison}
\end{table}

\paragraph{\textit{Finding 6: }\sys achieves the best performance but incurs higher computational overhead due to its search-based paradigm.}
We further evaluate the computational overhead of our method, \sys, in comparison with a set of competitive baselines on the Musique dataset using the Qwen2.5-14B-Instruct model. 
The results are summarized in \autoref{tab:baseline_comparison}.
As shown, \sys achieves the highest EM and F1 scores, demonstrating the effectiveness of our search-based reasoning strategy. 
However, this comes at a higher computational cost, primarily due to the enlarged search space and multi-step reasoning process.
In practical applications, the trade-off between performance and reasoning time should be contextualized by the nature of the task at hand. 
In high-stakes domains such as finance, healthcare, and law, sacrificing additional computational time for improved accuracy is often a worthwhile investment.

\subsection{Ablation Studies}

\subsubsection{Risk Assessment}
\begin{table*}[h]
\centering
\small 
\renewcommand{\arraystretch}{1.2} 
\setlength{\tabcolsep}{3pt} 
\begin{tabular}{c>{\centering\arraybackslash}p{1.2cm} >{\centering\arraybackslash}p{1.2cm}
                                  >{\centering\arraybackslash}p{1.2cm} >{\centering\arraybackslash}p{1.2cm}
                                  >{\centering\arraybackslash}p{1.2cm} >{\centering\arraybackslash}p{1.2cm}
                                  >{\centering\arraybackslash}p{1.2cm} >{\centering\arraybackslash}p{1.2cm}}
\toprule
\multirow{2}{*}{\textbf{Method}} & \multicolumn{2}{c}{\textbf{HotpotQA}} & \multicolumn{2}{c}{\textbf{2Wiki}} & \multicolumn{2}{c}{\textbf{MusiQue}} & \multicolumn{2}{c}{\textbf{Average}} \\ 
\cmidrule(lr){2-3} \cmidrule(lr){4-5} \cmidrule(lr){6-7} \cmidrule(lr){8-9}  
                        & \textbf{EM}  & \textbf{F1}  & \textbf{EM}  & \textbf{F1}  & \textbf{EM}  & \textbf{F1}  & \textbf{EM}  & \textbf{F1}  \\ 
\midrule
\textbf{MCTS (R-V function)}    & \textbf{73.50}  & \textbf{75.39}  & \textbf{56.50}  & \textbf{62.61}  & \textbf{40.50}  & \textbf{65.98}  & \textbf{56.83}  & \textbf{67.96}  \\  
MCTS (vanilla)                  & 71.00  & 73.58  & 48.50  & 60.47  & 34.50  & 63.34  & 51.33  & 65.80  \\  
MCTS (LLM-as-verifier)           & 69.00  & 73.85  & 53.50  & 60.98  & 34.00  & 64.30  & 52.17  & 66.38  \\  
\bottomrule
\end{tabular}
\caption{\textbf{Ablation on the Risk-Value function.} The incorporation of the Risk-Value function resulted in improvements across all datasets. The Risk-Value function effectively assesses reasoning states, guiding the tree search process; however, pretrained LLMs are not adequate as verifiers.}
\label{tab:ablation}
\end{table*}
\paragraph{\textit{Finding 7: }The Risk-Value function effectively risk-assess reasoning states to guide the tree search process.}
We conducted ablation studies to evaluate the effectiveness of the Risk-Value (R-V) function in guiding Monte Carlo Tree Search (MCTS).
\autoref{tab:ablation} present experimental results compared to vanilla MCTS and MCTS with LLM-as-Judge\footnote{To ensure experimental fairness, we employed the same LLM (policy models) for methods involving LLM-as-verifier.} baselines.
The incorporation of the R-V function resulted in improvements across all datasets.
Specifically, it achieved an average relative performance gain of 10.71\% over the vanilla MCTS baseline.
The function's impact was even more pronounced in more challenging tasks, with improvements reaching up to 17.39\% on MusiQue.
This demonstrates the function's capacity to better navigate and prioritize lower-risk paths, ensuring more efficient exploration and exploitation within the search space.
In comparison, MCTS with LLM-as-Verifier showed marginal improvements over the vanilla approach.
While pretrained LLMs can provide meaningful context during verification, they are not specifically tuned for evaluating the quality of reasoning states in a dynamic environment.
This suggests that pretrained LLMs are insufficient as standalone verifiers in path planning, underscoring the critical role of specialized functions like Risk-Value in guiding the search process.

\subsubsection{Iterations of MCTS}
\begin{figure}[h]
    \centering
    \includegraphics[width=1\linewidth]{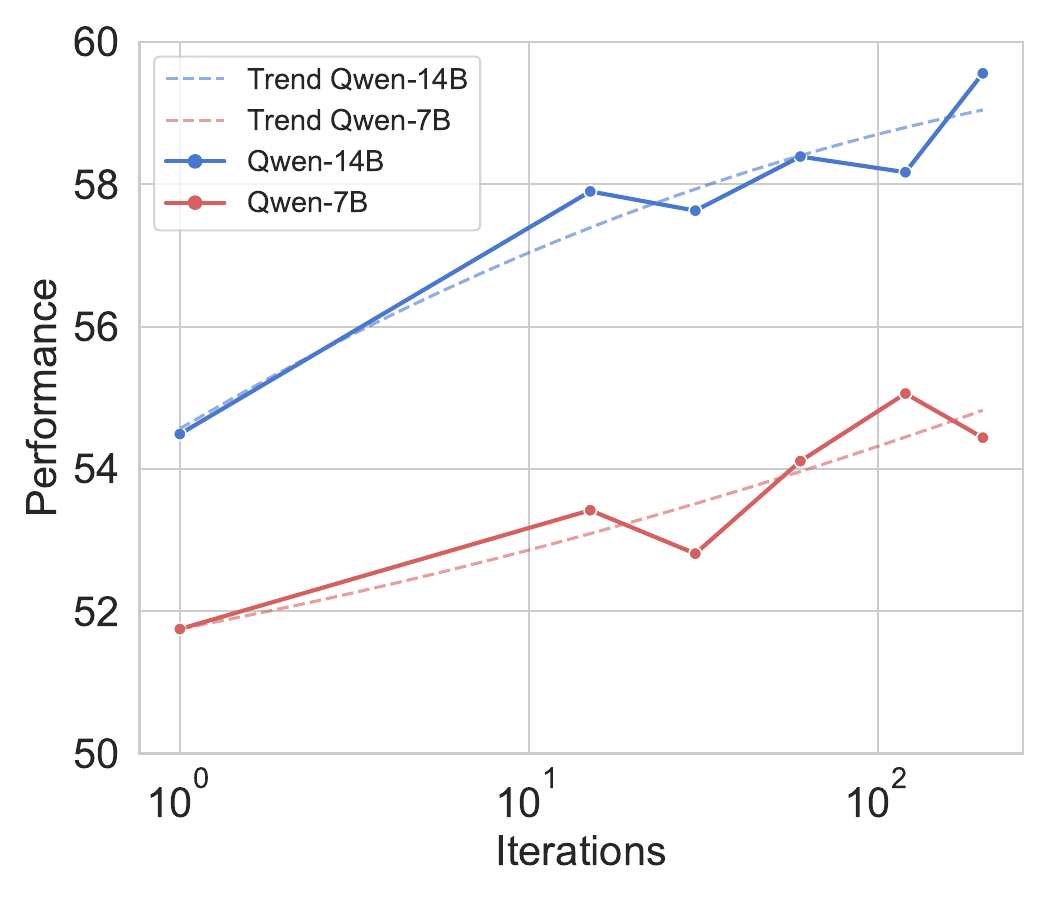}
    \caption{
    \textbf{Ablation on iterations.} We evaluate performance using the average of EM accuracy and F1 score.
    }
    \label{fig:iter}
\end{figure}

\paragraph{\textit{Finding 8: }\sys with dynamic risk assessment achieves near-optimal solutions with a relatively low inference cost.}
We conduct empirical experiments on the trade-off between inference cost and solution quality.
Specifically, we examined the performance of \sys as the number of MCTS iterations increased from 1 to 200.
\autoref{fig:iter} illustrates the efficiency of \sys in reaching near-optimal solutions with a relatively low inference cost.
As the number of MCTS iterations increases, the performance improves, but the rate of improvement diminishes with higher iteration counts.
At the initial stages of the search process (from 1 to 30 iterations), the performance shows rapid improvements.
The initial exploration of potential moves encourages the model's better understanding of the search space.
The subsequent increase in the number of additional paths during this phase contributes meaningfully to the quality of the solution.
Beyond 30 iterations, the improvements in performance begin to level off.
For instance, between 30 and 60 iterations, the performance increases slightly by a relative 1.32\%, whereas during the initial phase, the increase was 5.76\%.
This suggests that further exploration of the search space yields diminishing returns as the algorithm begins to converge toward the optimal decision.
This phenomenon can be attributed to that \sys provides dynamic step-level Risk-Value, without requiring the model to wait for outcome verification to guide the next iteration.
%
The Risk-Value function efficiently narrows down the promising branches at a rapid pace.
The performance stabilizes after approximately 100 iterations.
Further iterations may lead to incremental improvements but are less impactful in a highly explored search space.
%
%

\section{Related Works}

\paragraph{Test-Time Compute.}
Scaling test-time compute has become a popular topic recently in many research efforts~\citep{DBLP:journals/corr/abs-2407-21787, DBLP:journals/corr/abs-2408-03314}, aiming to shift the reasoning paradigm of LLMs from the fast but error-prone system 1 thinking to the more deliberate system 2 thinking~\citep{kahneman2011thinking,DBLP:journals/corr/abs-2408-03314}.
Prior works mainly include prompt-based~\cite{DBLP:journals/csur/YuZTW24}, search-based~\cite{DBLP:conf/emnlp/HaoGMHWWH23}, and learning-based~\cite{OpenAI-o1,deepseek-r1} methods.
Prompt-based reasoning utilize chain-of-thought (CoT)~\citep{DBLP:conf/nips/Wei0SBIXCLZ22} prompting to break down complex reasoning into sub-steps, gradually approaching a more accurate final answer through the generation of additional intermediate tokens~\cite{DBLP:conf/acl/ZhaoLJQB23,DBLP:journals/corr/abs-2411-19443, DBLP:conf/iclr/LiZCDJPB24}.
Search-based reasoning allows LLMs to improve performance by generating multiple samples, and tree-based methods have further integrated planning and exploration \citep{DBLP:conf/nips/YaoYZS00N23, DBLP:conf/aaai/BestaBKGPGGLNNH24, DBLP:conf/emnlp/HaoGMHWWH23}. 
While the multiple and redundant rollouts significantly burden the inference spend, verifiers for solution-selection is essential for ensuring efficiency.
Learning-based methods aim to inject the deep reasoning patterns of large and complicated models into smaller models through post-training~\citep{OpenAI-o1,deepseek-r1}.

\paragraph{Retrieval-Augmented Generation.}
Retrieval-Augmented Generation (RAG) merges the intrinsic knowledge of LLMs with a vast, dynamic repository of external databases, mitigating the issues of language model hallucination and outdated knowledge to some extent \citep{DBLP:conf/nips/LewisPPPKGKLYR020, DBLP:journals/corr/abs-2312-10997}.
Recent studies \citep{DBLP:conf/emnlp/WangYW23, DBLP:conf/emnlp/PressZMSSL23, DBLP:journals/corr/abs-2411-19443, DBLP:conf/acl/ZhaoLJQB23} have proposed some prompting-based strategies for LLMs to better harness the potential of RAG, essentially integrating it into the intermediate reasoning process (\eg chain-of-thought (CoT) \citep{DBLP:conf/nips/Wei0SBIXCLZ22}). 
In these methods, the interaction between LLMs and retrieval actions breaks down the reasoning process into discontinuous smaller steps, which helps produce more authentic intermediate results and reduces the instability inherent in autoregressive token generation.
\section{Conclusion}


In this work, we introduce \sys, a novel framework that addresses the challenges of error propagation and verification bottleneck in open-ended, knowledge-intensive, and complex reasoning tasks. By integrating Monte Carlo Tree Search with risk-adaptive exploration, \sys enables dynamic and effective reasoning through iterative problem-decomposition and retrieval-then-reasoning steps. Our experiments demonstrate that \sys outperforms a wide range of state-of-the-art methods, and also surpasses the performance of the latest learning-based large reasoning models (LRMs) equipped with RAG. These results highlight the strong potential of \sys in advancing open-ended, knowledge-intensive, and complex reasoning tasks across various real-world applications.
\clearpage
\section*{Limitations}
\label{sec:lim}

Although our method \sys demonstrates strong performance for knowledge-intensive, and complex reasoning tasks, several limitations remain open for improvement.
Our experiments are currently confined to multi-hop question-answering (QA) tasks.
The applicability of \sys to other reasoning tasks, such as mathematical problem-solving, code generation, or complex decision-making, remains to be explored. 
Extending our method to a broader range of reasoning tasks is an important direction for future work.
There is also a need for more systematically designed prompts to ensure generalization and robustness across diverse scenarios.
Moreover, existing search-based paradigms mainly rely on post-trained reward models for verification. 
These models are trained on specially annotated data to learn how to score reasoning states. 
While this approach has shown progress in mathematics (a relatively closed domain)~\cite{DBLP:conf/iclr/LightmanKBEBLLS24, DBLP:journals/corr/abs-2406-06592, DBLP:conf/acl/WangLSXDLCWS24}, static, post-trained reward models struggle with open-ended, knowledge-intensive tasks, where external knowledge is dynamically involved. 
Such models cannot accurately assess new knowledge unseen during post-training, limiting their generalization.
We argue that a generalizable design of rewards is crucial for future developments. 
Our work presents an initial attempt in this direction, and we will further explore this line of research.
Last but not least, the search space for specific questions is predefined, lacking the ability to adapt efficiently to varying reasoning complexity or knowledge density. 
This points to an open challenge in the study of efficient reasoning: how to achieve an effective trade-off between broad and deep search, and how to reduce redundancy in the reasoning process.
We leave this as an avenue for future work.

\section*{Acknowledgement}
\label{sec:ack}
We thank all the anonymous reviewers and area chair for their valuable feedback throughout the
review process.
This work is supported by Shanghai Artificial Intelligence Laboratory.


\bibliography{anthology,src/custom}

\begin{thebibliography}{53}
\providecommand{\natexlab}[1]{#1}

\bibitem[{Amirizaniani et~al.(2024)Amirizaniani, Martin, Sivachenko, Mashhadi, and Shah}]{DBLP:conf/cikm/AmirizanianiMSM24}
Maryam Amirizaniani, Elias Martin, Maryna Sivachenko, Afra Mashhadi, and Chirag Shah. 2024.
\newblock \href {https://doi.org/10.1145/3627673.3679832} {Can llms reason like humans? assessing theory of mind reasoning in llms for open-ended questions}.
\newblock In \emph{Proceedings of the 33rd {ACM} International Conference on Information and Knowledge Management, {CIKM} 2024, Boise, ID, USA, October 21-25, 2024}, pages 34--44. {ACM}.

\bibitem[{Besta et~al.(2024)Besta, Blach, Kubicek, Gerstenberger, Podstawski, Gianinazzi, Gajda, Lehmann, Niewiadomski, Nyczyk, and Hoefler}]{DBLP:conf/aaai/BestaBKGPGGLNNH24}
Maciej Besta, Nils Blach, Ales Kubicek, Robert Gerstenberger, Michal Podstawski, Lukas Gianinazzi, Joanna Gajda, Tomasz Lehmann, Hubert Niewiadomski, Piotr Nyczyk, and Torsten Hoefler. 2024.
\newblock \href {https://doi.org/10.1609/AAAI.V38I16.29720} {Graph of thoughts: Solving elaborate problems with large language models}.
\newblock In \emph{Thirty-Eighth {AAAI} Conference on Artificial Intelligence, {AAAI} 2024, Thirty-Sixth Conference on Innovative Applications of Artificial Intelligence, {IAAI} 2024, Fourteenth Symposium on Educational Advances in Artificial Intelligence, {EAAI} 2014, February 20-27, 2024, Vancouver, Canada}, pages 17682--17690. {AAAI} Press.

\bibitem[{Brown et~al.(2024)Brown, Juravsky, Ehrlich, Clark, Le, R{\'{e}}, and Mirhoseini}]{DBLP:journals/corr/abs-2407-21787}
Bradley C.~A. Brown, Jordan Juravsky, Ryan~Saul Ehrlich, Ronald Clark, Quoc~V. Le, Christopher R{\'{e}}, and Azalia Mirhoseini. 2024.
\newblock \href {https://doi.org/10.48550/ARXIV.2407.21787} {Large language monkeys: Scaling inference compute with repeated sampling}.
\newblock \emph{CoRR}, abs/2407.21787.

\bibitem[{Bubeck et~al.(2023)Bubeck, Chandrasekaran, Eldan, Gehrke, Horvitz, Kamar, Lee, Lee, Li, Lundberg, Nori, Palangi, Ribeiro, and Zhang}]{DBLP:journals/corr/abs-2303-12712}
S{\'{e}}bastien Bubeck, Varun Chandrasekaran, Ronen Eldan, Johannes Gehrke, Eric Horvitz, Ece Kamar, Peter Lee, Yin~Tat Lee, Yuanzhi Li, Scott~M. Lundberg, Harsha Nori, Hamid Palangi, Marco~T{\'{u}}lio Ribeiro, and Yi~Zhang. 2023.
\newblock \href {https://doi.org/10.48550/ARXIV.2303.12712} {Sparks of artificial general intelligence: Early experiments with {GPT-4}}.
\newblock \emph{CoRR}, abs/2303.12712.

\bibitem[{Cobbe et~al.(2021)Cobbe, Kosaraju, Bavarian, Chen, Jun, Kaiser, Plappert, Tworek, Hilton, Nakano, Hesse, and Schulman}]{DBLP:journals/corr/abs-2110-14168}
Karl Cobbe, Vineet Kosaraju, Mohammad Bavarian, Mark Chen, Heewoo Jun, Lukasz Kaiser, Matthias Plappert, Jerry Tworek, Jacob Hilton, Reiichiro Nakano, Christopher Hesse, and John Schulman. 2021.
\newblock \href {https://arxiv.org/abs/2110.14168} {Training verifiers to solve math word problems}.
\newblock \emph{CoRR}, abs/2110.14168.

\bibitem[{DeepSeek-AI et~al.(2025)DeepSeek-AI, Guo, Yang, Zhang, Song, Zhang, Xu, Zhu, Ma, Wang, Bi, Zhang, Yu, Wu, Wu, Gou, Shao, Li, Gao, Liu, Xue, Wang, Wu, Feng, Lu, Zhao, Deng, Zhang, Ruan, Dai, Chen, Ji, Li, Lin, Dai, Luo, Hao, Chen, Li, Zhang, Bao, Xu, Wang, Ding, Xin, Gao, Qu, Li, Guo, Li, Wang, Chen, Yuan, Qiu, Li, Cai, Ni, Liang, Chen, Dong, Hu, Gao, Guan, Huang, Yu, Wang, Zhang, Zhao, Wang, Zhang, Xu, Xia, Zhang, Zhang, Tang, Li, Wang, Li, Tian, Huang, Zhang, Wang, Chen, Du, Ge, Zhang, Pan, Wang, Chen, Jin, Chen, Lu, Zhou, Chen, Ye, Wang, Yu, Zhou, Pan, Li, Zhou, Wu, Ye, Yun, Pei, Sun, Wang, Zeng, Zhao, Liu, Liang, Gao, Yu, Zhang, Xiao, An, Liu, Wang, Chen, Nie, Cheng, Liu, Xie, Liu, Yang, Li, Su, Lin, Li, Jin, Shen, Chen, Sun, Wang, Song, Zhou, Wang, Shan, Li, Wang, Wei, Zhang, Xu, Li, Zhao, Sun, Wang, Yu, Zhang, Shi, Xiong, He, Piao, Wang, Tan, Ma, Liu, Guo, Ou, Wang, Gong, Zou, He, Xiong, Luo, You, Liu, Zhou, Zhu, Xu, Huang, Li, Zheng, Zhu, Ma, Tang, Zha, Yan, Ren, Ren, Sha, Fu, Xu, Xie, Zhang,
  Hao, Ma, Yan, Wu, Gu, Zhu, Liu, Li, Xie, Song, Pan, Huang, Xu, Zhang, and Zhang}]{deepseek-r1}
DeepSeek-AI, Daya Guo, Dejian Yang, Haowei Zhang, Junxiao Song, Ruoyu Zhang, Runxin Xu, Qihao Zhu, Shirong Ma, Peiyi Wang, Xiao Bi, Xiaokang Zhang, Xingkai Yu, Yu~Wu, Z.~F. Wu, Zhibin Gou, Zhihong Shao, Zhuoshu Li, Ziyi Gao, and 181 others. 2025.
\newblock \href {https://arxiv.org/abs/2501.12948} {Deepseek-r1: Incentivizing reasoning capability in llms via reinforcement learning}.

\bibitem[{Dubey et~al.(2024)Dubey, Jauhri, Pandey, Kadian, Al{-}Dahle, Letman, Mathur, Schelten, Yang, Fan, Goyal, Hartshorn, Yang, Mitra, Sravankumar, Korenev, Hinsvark, Rao, Zhang, Rodriguez, Gregerson, Spataru, Rozi{\`{e}}re, Biron, Tang, Chern, Caucheteux, Nayak, Bi, Marra, McConnell, Keller, Touret, Wu, Wong, Ferrer, Nikolaidis, Allonsius, Song, Pintz, Livshits, Esiobu, Choudhary, Mahajan, Garcia{-}Olano, Perino, Hupkes, Lakomkin, AlBadawy, Lobanova, Dinan, Smith, Radenovic, Zhang, Synnaeve, Lee, Anderson, Nail, Mialon, Pang, Cucurell, Nguyen, Korevaar, Xu, Touvron, Zarov, Ibarra, Kloumann, Misra, Evtimov, Copet, Lee, Geffert, Vranes, Park, Mahadeokar, Shah, van~der Linde, Billock, Hong, Lee, Fu, Chi, Huang, Liu, Wang, Yu, Bitton, Spisak, Park, Rocca, Johnstun, Saxe, Jia, Alwala, Upasani, Plawiak, Li, Heafield, Stone, and et~al.}]{llama}
Abhimanyu Dubey, Abhinav Jauhri, Abhinav Pandey, Abhishek Kadian, Ahmad Al{-}Dahle, Aiesha Letman, Akhil Mathur, Alan Schelten, Amy Yang, Angela Fan, Anirudh Goyal, Anthony Hartshorn, Aobo Yang, Archi Mitra, Archie Sravankumar, Artem Korenev, Arthur Hinsvark, Arun Rao, Aston Zhang, and 82 others. 2024.
\newblock \href {https://doi.org/10.48550/ARXIV.2407.21783} {The llama 3 herd of models}.
\newblock \emph{CoRR}, abs/2407.21783.

\bibitem[{Feng et~al.(2025)Feng, Hao, Zhang, Song, and Wang}]{feng2025airrag}
Wenfeng Feng, Chuzhan Hao, Yuewei Zhang, Jingyi Song, and Hao Wang. 2025.
\newblock \href {https://arxiv.org/abs/2501.10053} {Airrag: Activating intrinsic reasoning for retrieval augmented generation via tree-based search}.
\newblock \emph{Preprint}, arXiv:2501.10053.

\bibitem[{Gao et~al.(2023)Gao, Xiong, Gao, Jia, Pan, Bi, Dai, Sun, Guo, Wang, and Wang}]{DBLP:journals/corr/abs-2312-10997}
Yunfan Gao, Yun Xiong, Xinyu Gao, Kangxiang Jia, Jinliu Pan, Yuxi Bi, Yi~Dai, Jiawei Sun, Qianyu Guo, Meng Wang, and Haofen Wang. 2023.
\newblock \href {https://doi.org/10.48550/ARXIV.2312.10997} {Retrieval-augmented generation for large language models: {A} survey}.
\newblock \emph{CoRR}, abs/2312.10997.

\bibitem[{Hao et~al.(2023)Hao, Gu, Ma, Hong, Wang, Wang, and Hu}]{DBLP:conf/emnlp/HaoGMHWWH23}
Shibo Hao, Yi~Gu, Haodi Ma, Joshua~Jiahua Hong, Zhen Wang, Daisy~Zhe Wang, and Zhiting Hu. 2023.
\newblock \href {https://doi.org/10.18653/V1/2023.EMNLP-MAIN.507} {Reasoning with language model is planning with world model}.
\newblock In \emph{Proceedings of the 2023 Conference on Empirical Methods in Natural Language Processing, {EMNLP} 2023, Singapore, December 6-10, 2023}, pages 8154--8173. Association for Computational Linguistics.

\bibitem[{Ho et~al.(2020)Ho, Nguyen, Sugawara, and Aizawa}]{DBLP:conf/coling/HoNSA20}
Xanh Ho, Anh{-}Khoa~Duong Nguyen, Saku Sugawara, and Akiko Aizawa. 2020.
\newblock \href {https://doi.org/10.18653/V1/2020.COLING-MAIN.580} {Constructing {A} multi-hop {QA} dataset for comprehensive evaluation of reasoning steps}.
\newblock In \emph{Proceedings of the 28th International Conference on Computational Linguistics, {COLING} 2020, Barcelona, Spain (Online), December 8-13, 2020}, pages 6609--6625. International Committee on Computational Linguistics.

\bibitem[{Jiang et~al.(2024)Jiang, Chen, Li, Ren, Wang, Zhao, Song, and Zhang}]{DBLP:journals/corr/abs-2412-12881}
Jinhao Jiang, Jiayi Chen, Junyi Li, Ruiyang Ren, Shijie Wang, Wayne~Xin Zhao, Yang Song, and Tao Zhang. 2024.
\newblock \href {https://doi.org/10.48550/ARXIV.2412.12881} {Rag-star: Enhancing deliberative reasoning with retrieval augmented verification and refinement}.
\newblock \emph{CoRR}, abs/2412.12881.

\bibitem[{Kahneman(2011)}]{kahneman2011thinking}
Daniel Kahneman. 2011.
\newblock \emph{Thinking, fast and slow}.
\newblock Farrar, Straus and Giroux.

\bibitem[{Kocsis and Szepesv{\'{a}}ri(2006)}]{DBLP:conf/ecml/KocsisS06}
Levente Kocsis and Csaba Szepesv{\'{a}}ri. 2006.
\newblock \href {https://doi.org/10.1007/11871842\_29} {Bandit based monte-carlo planning}.
\newblock In \emph{Machine Learning: {ECML} 2006, 17th European Conference on Machine Learning, Berlin, Germany, September 18-22, 2006, Proceedings}, volume 4212 of \emph{Lecture Notes in Computer Science}, pages 282--293. Springer.

\bibitem[{Lafferty and Zhai(2001)}]{DBLP:conf/sigir/LaffertyZ01}
John~D. Lafferty and ChengXiang Zhai. 2001.
\newblock \href {https://doi.org/10.1145/383952.383970} {Document language models, query models, and risk minimization for information retrieval}.
\newblock In \emph{{SIGIR} 2001: Proceedings of the 24th Annual International {ACM} {SIGIR} Conference on Research and Development in Information Retrieval, September 9-13, 2001, New Orleans, Louisiana, {USA}}, pages 111--119. {ACM}.

\bibitem[{Lewis et~al.(2020)Lewis, Perez, Piktus, Petroni, Karpukhin, Goyal, K{\"{u}}ttler, Lewis, Yih, Rockt{\"{a}}schel, Riedel, and Kiela}]{DBLP:conf/nips/LewisPPPKGKLYR020}
Patrick S.~H. Lewis, Ethan Perez, Aleksandra Piktus, Fabio Petroni, Vladimir Karpukhin, Naman Goyal, Heinrich K{\"{u}}ttler, Mike Lewis, Wen{-}tau Yih, Tim Rockt{\"{a}}schel, Sebastian Riedel, and Douwe Kiela. 2020.
\newblock \href {https://proceedings.neurips.cc/paper/2020/hash/6b493230205f780e1bc26945df7481e5-Abstract.html} {Retrieval-augmented generation for knowledge-intensive {NLP} tasks}.
\newblock In \emph{Advances in Neural Information Processing Systems 33: Annual Conference on Neural Information Processing Systems 2020, NeurIPS 2020, December 6-12, 2020, virtual}.

\bibitem[{Li et~al.(2024)Li, Zhao, Chia, Ding, Joty, Poria, and Bing}]{DBLP:conf/iclr/LiZCDJPB24}
Xingxuan Li, Ruochen Zhao, Yew~Ken Chia, Bosheng Ding, Shafiq Joty, Soujanya Poria, and Lidong Bing. 2024.
\newblock \href {https://openreview.net/forum?id=cPgh4gWZlz} {Chain-of-knowledge: Grounding large language models via dynamic knowledge adapting over heterogeneous sources}.
\newblock In \emph{The Twelfth International Conference on Learning Representations, {ICLR} 2024, Vienna, Austria, May 7-11, 2024}. OpenReview.net.

\bibitem[{Li et~al.(2023)Li, Wang, Ding, and Chen}]{DBLP:conf/icaif/LiWDC23}
Yinheng Li, Shaofei Wang, Han Ding, and Hang Chen. 2023.
\newblock \href {https://doi.org/10.1145/3604237.3626869} {Large language models in finance: {A} survey}.
\newblock In \emph{4th {ACM} International Conference on {AI} in Finance, {ICAIF} 2023, Brooklyn, NY, USA, November 27-29, 2023}, pages 374--382. {ACM}.

\bibitem[{Lightman et~al.(2024)Lightman, Kosaraju, Burda, Edwards, Baker, Lee, Leike, Schulman, Sutskever, and Cobbe}]{DBLP:conf/iclr/LightmanKBEBLLS24}
Hunter Lightman, Vineet Kosaraju, Yuri Burda, Harrison Edwards, Bowen Baker, Teddy Lee, Jan Leike, John Schulman, Ilya Sutskever, and Karl Cobbe. 2024.
\newblock \href {https://openreview.net/forum?id=v8L0pN6EOi} {Let's verify step by step}.
\newblock In \emph{The Twelfth International Conference on Learning Representations, {ICLR} 2024, Vienna, Austria, May 7-11, 2024}. OpenReview.net.

\bibitem[{Liu et~al.(2024)Liu, Lin, and Liu}]{DBLP:journals/corr/abs-2410-02338}
Jingyu Liu, Jiaen Lin, and Yong Liu. 2024.
\newblock \href {https://doi.org/10.48550/ARXIV.2410.02338} {How much can {RAG} help the reasoning of llm?}
\newblock \emph{CoRR}, abs/2410.02338.

\bibitem[{Luo et~al.(2024)Luo, Liu, Liu, Phatale, Lara, Li, Shu, Zhu, Meng, Sun, and Rastogi}]{DBLP:journals/corr/abs-2406-06592}
Liangchen Luo, Yinxiao Liu, Rosanne Liu, Samrat Phatale, Harsh Lara, Yunxuan Li, Lei Shu, Yun Zhu, Lei Meng, Jiao Sun, and Abhinav Rastogi. 2024.
\newblock \href {https://doi.org/10.48550/ARXIV.2406.06592} {Improve mathematical reasoning in language models by automated process supervision}.
\newblock \emph{CoRR}, abs/2406.06592.

\bibitem[{OpenAI(2023)}]{DBLP:journals/corr/abs-2303-08774}
OpenAI. 2023.
\newblock \href {https://doi.org/10.48550/arXiv.2303.08774} {{GPT-4} technical report}.

\bibitem[{OpenAI(2024)}]{OpenAI-o1}
OpenAI. 2024.
\newblock \href {https://openai.com/index/learning-to-reason-with-llms/} {Learning to reason with llms}.

\bibitem[{OpenAI(2025)}]{OpenAI-deep-research}
OpenAI. 2025.
\newblock \href {https://openai.com/index/introducing-deep-research/} {Introducing deep research}.

\bibitem[{Ponte and Croft(1998)}]{DBLP:conf/sigir/PonteC98}
Jay~M. Ponte and W.~Bruce Croft. 1998.
\newblock \href {https://doi.org/10.1145/290941.291008} {A language modeling approach to information retrieval}.
\newblock In \emph{{SIGIR} '98: Proceedings of the 21st Annual International {ACM} {SIGIR} Conference on Research and Development in Information Retrieval, August 24-28 1998, Melbourne, Australia}, pages 275--281. {ACM}.

\bibitem[{Press et~al.(2023)Press, Zhang, Min, Schmidt, Smith, and Lewis}]{DBLP:conf/emnlp/PressZMSSL23}
Ofir Press, Muru Zhang, Sewon Min, Ludwig Schmidt, Noah~A. Smith, and Mike Lewis. 2023.
\newblock \href {https://doi.org/10.18653/V1/2023.FINDINGS-EMNLP.378} {Measuring and narrowing the compositionality gap in language models}.
\newblock In \emph{Findings of the Association for Computational Linguistics: {EMNLP} 2023, Singapore, December 6-10, 2023}, pages 5687--5711. Association for Computational Linguistics.

\bibitem[{Rajani et~al.(2019)Rajani, McCann, Xiong, and Socher}]{DBLP:conf/acl/RajaniMXS19}
Nazneen~Fatema Rajani, Bryan McCann, Caiming Xiong, and Richard Socher. 2019.
\newblock \href {https://doi.org/10.18653/V1/P19-1487} {Explain yourself! leveraging language models for commonsense reasoning}.
\newblock In \emph{Proceedings of the 57th Conference of the Association for Computational Linguistics, {ACL} 2019, Florence, Italy, July 28- August 2, 2019, Volume 1: Long Papers}, pages 4932--4942. Association for Computational Linguistics.

\bibitem[{Sachan et~al.(2022)Sachan, Lewis, Joshi, Aghajanyan, Yih, Pineau, and Zettlemoyer}]{DBLP:conf/emnlp/SachanLJAYPZ22}
Devendra~Singh Sachan, Mike Lewis, Mandar Joshi, Armen Aghajanyan, Wen{-}tau Yih, Joelle Pineau, and Luke Zettlemoyer. 2022.
\newblock \href {https://doi.org/10.18653/V1/2022.EMNLP-MAIN.249} {Improving passage retrieval with zero-shot question generation}.
\newblock In \emph{Proceedings of the 2022 Conference on Empirical Methods in Natural Language Processing, {EMNLP} 2022, Abu Dhabi, United Arab Emirates, December 7-11, 2022}, pages 3781--3797. Association for Computational Linguistics.

\bibitem[{Setlur et~al.(2024)Setlur, Nagpal, Fisch, Geng, Eisenstein, Agarwal, Agarwal, Berant, and Kumar}]{DBLP:journals/corr/abs-2410-08146}
Amrith Setlur, Chirag Nagpal, Adam Fisch, Xinyang Geng, Jacob Eisenstein, Rishabh Agarwal, Alekh Agarwal, Jonathan Berant, and Aviral Kumar. 2024.
\newblock \href {https://doi.org/10.48550/ARXIV.2410.08146} {Rewarding progress: Scaling automated process verifiers for {LLM} reasoning}.
\newblock \emph{CoRR}, abs/2410.08146.

\bibitem[{Snell et~al.(2024)Snell, Lee, Xu, and Kumar}]{DBLP:journals/corr/abs-2408-03314}
Charlie Snell, Jaehoon Lee, Kelvin Xu, and Aviral Kumar. 2024.
\newblock \href {https://doi.org/10.48550/ARXIV.2408.03314} {Scaling {LLM} test-time compute optimally can be more effective than scaling model parameters}.
\newblock \emph{CoRR}, abs/2408.03314.

\bibitem[{Stechly et~al.(2024)Stechly, Valmeekam, and Kambhampati}]{DBLP:journals/corr/abs-2402-08115}
Kaya Stechly, Karthik Valmeekam, and Subbarao Kambhampati. 2024.
\newblock \href {https://doi.org/10.48550/ARXIV.2402.08115} {On the self-verification limitations of large language models on reasoning and planning tasks}.
\newblock \emph{CoRR}, abs/2402.08115.

\bibitem[{Taylor et~al.(2022)Taylor, Kardas, Cucurull, Scialom, Hartshorn, Saravia, Poulton, Kerkez, and Stojnic}]{DBLP:journals/corr/abs-2211-09085}
Ross Taylor, Marcin Kardas, Guillem Cucurull, Thomas Scialom, Anthony Hartshorn, Elvis Saravia, Andrew Poulton, Viktor Kerkez, and Robert Stojnic. 2022.
\newblock \href {https://doi.org/10.48550/ARXIV.2211.09085} {Galactica: {A} large language model for science}.
\newblock \emph{CoRR}, abs/2211.09085.

\bibitem[{Team(2024)}]{qwen2.5}
Qwen Team. 2024.
\newblock \href {https://qwenlm.github.io/blog/qwen2.5/} {Qwen2.5: A party of foundation models}.

\bibitem[{Thirunavukarasu et~al.(2023)Thirunavukarasu, Ting, Elangovan, Gutierrez, Tan, and Ting}]{thirunavukarasu2023large}
Arun~James Thirunavukarasu, Darren Shu~Jeng Ting, Kabilan Elangovan, Laura Gutierrez, Ting~Fang Tan, and Daniel Shu~Wei Ting. 2023.
\newblock \href {https://doi.org/10.1038/s41591-023-02448-8} {Large language models in medicine}.
\newblock \emph{Nature Medicine}, 29(8):1930--1940.

\bibitem[{Trivedi et~al.(2022)Trivedi, Balasubramanian, Khot, and Sabharwal}]{DBLP:journals/tacl/TrivediBKS22}
Harsh Trivedi, Niranjan Balasubramanian, Tushar Khot, and Ashish Sabharwal. 2022.
\newblock \href {https://doi.org/10.1162/TACL\_A\_00475} {Musique: Multihop questions via single-hop question composition}.
\newblock \emph{Trans. Assoc. Comput. Linguistics}, 10:539--554.

\bibitem[{Valmeekam et~al.(2022)Valmeekam, Hernandez, Sreedharan, and Kambhampati}]{DBLP:journals/corr/abs-2206-10498}
Karthik Valmeekam, Alberto~Olmo Hernandez, Sarath Sreedharan, and Subbarao Kambhampati. 2022.
\newblock \href {https://doi.org/10.48550/ARXIV.2206.10498} {Large language models still can't plan {(A} benchmark for llms on planning and reasoning about change)}.
\newblock \emph{CoRR}, abs/2206.10498.

\bibitem[{Wang et~al.(2023{\natexlab{a}})Wang, Yang, and Wei}]{DBLP:conf/emnlp/WangYW23}
Liang Wang, Nan Yang, and Furu Wei. 2023{\natexlab{a}}.
\newblock \href {https://doi.org/10.18653/V1/2023.EMNLP-MAIN.585} {Query2doc: Query expansion with large language models}.
\newblock In \emph{Proceedings of the 2023 Conference on Empirical Methods in Natural Language Processing, {EMNLP} 2023, Singapore, December 6-10, 2023}, pages 9414--9423. Association for Computational Linguistics.

\bibitem[{Wang et~al.(2024)Wang, Li, Shao, Xu, Dai, Li, Chen, Wu, and Sui}]{DBLP:conf/acl/WangLSXDLCWS24}
Peiyi Wang, Lei Li, Zhihong Shao, Runxin Xu, Damai Dai, Yifei Li, Deli Chen, Yu~Wu, and Zhifang Sui. 2024.
\newblock \href {https://doi.org/10.18653/V1/2024.ACL-LONG.510} {Math-shepherd: Verify and reinforce llms step-by-step without human annotations}.
\newblock In \emph{Proceedings of the 62nd Annual Meeting of the Association for Computational Linguistics (Volume 1: Long Papers), {ACL} 2024, Bangkok, Thailand, August 11-16, 2024}, pages 9426--9439. Association for Computational Linguistics.

\bibitem[{Wang et~al.(2023{\natexlab{b}})Wang, Wei, Schuurmans, Le, Chi, Narang, Chowdhery, and Zhou}]{DBLP:conf/iclr/0002WSLCNCZ23}
Xuezhi Wang, Jason Wei, Dale Schuurmans, Quoc~V. Le, Ed~H. Chi, Sharan Narang, Aakanksha Chowdhery, and Denny Zhou. 2023{\natexlab{b}}.
\newblock \href {https://openreview.net/forum?id=1PL1NIMMrw} {Self-consistency improves chain of thought reasoning in language models}.
\newblock In \emph{The Eleventh International Conference on Learning Representations, {ICLR} 2023, Kigali, Rwanda, May 1-5, 2023}. OpenReview.net.

\bibitem[{Wei et~al.(2022)Wei, Wang, Schuurmans, Bosma, Ichter, Xia, Chi, Le, and Zhou}]{DBLP:conf/nips/Wei0SBIXCLZ22}
Jason Wei, Xuezhi Wang, Dale Schuurmans, Maarten Bosma, Brian Ichter, Fei Xia, Ed~H. Chi, Quoc~V. Le, and Denny Zhou. 2022.
\newblock \href {http://papers.nips.cc/paper\_files/paper/2022/hash/9d5609613524ecf4f15af0f7b31abca4-Abstract-Conference.html} {Chain-of-thought prompting elicits reasoning in large language models}.
\newblock In \emph{Advances in Neural Information Processing Systems 35: Annual Conference on Neural Information Processing Systems 2022, NeurIPS 2022, New Orleans, LA, USA, November 28 - December 9, 2022}.

\bibitem[{Wies et~al.(2023)Wies, Levine, and Shashua}]{DBLP:conf/iclr/WiesLS23}
Noam Wies, Yoav Levine, and Amnon Shashua. 2023.
\newblock \href {https://openreview.net/forum?id=BrJATVZDWEH} {Sub-task decomposition enables learning in sequence to sequence tasks}.
\newblock In \emph{The Eleventh International Conference on Learning Representations, {ICLR} 2023, Kigali, Rwanda, May 1-5, 2023}. OpenReview.net.

\bibitem[{Xu et~al.(2025)Xu, Hao, Zong, Wang, Zhang, Wang, Lan, Gong, Ouyang, Meng, Shao, Yan, Yang, Song, Ren, Hu, Li, Feng, Gao, and Li}]{xu2025largereasoningmodelssurvey}
Fengli Xu, Qianyue Hao, Zefang Zong, Jingwei Wang, Yunke Zhang, Jingyi Wang, Xiaochong Lan, Jiahui Gong, Tianjian Ouyang, Fanjin Meng, Chenyang Shao, Yuwei Yan, Qinglong Yang, Yiwen Song, Sijian Ren, Xinyuan Hu, Yu~Li, Jie Feng, Chen Gao, and Yong Li. 2025.
\newblock \href {https://arxiv.org/abs/2501.09686} {Towards large reasoning models: A survey of reinforced reasoning with large language models}.

\bibitem[{Yang et~al.(2018)Yang, Qi, Zhang, Bengio, Cohen, Salakhutdinov, and Manning}]{DBLP:conf/emnlp/Yang0ZBCSM18}
Zhilin Yang, Peng Qi, Saizheng Zhang, Yoshua Bengio, William~W. Cohen, Ruslan Salakhutdinov, and Christopher~D. Manning. 2018.
\newblock \href {https://doi.org/10.18653/V1/D18-1259} {Hotpotqa: {A} dataset for diverse, explainable multi-hop question answering}.
\newblock In \emph{Proceedings of the 2018 Conference on Empirical Methods in Natural Language Processing, Brussels, Belgium, October 31 - November 4, 2018}, pages 2369--2380. Association for Computational Linguistics.

\bibitem[{Yao et~al.(2023)Yao, Yu, Zhao, Shafran, Griffiths, Cao, and Narasimhan}]{DBLP:conf/nips/YaoYZS00N23}
Shunyu Yao, Dian Yu, Jeffrey Zhao, Izhak Shafran, Tom Griffiths, Yuan Cao, and Karthik Narasimhan. 2023.
\newblock \href {http://papers.nips.cc/paper\_files/paper/2023/hash/271db9922b8d1f4dd7aaef84ed5ac703-Abstract-Conference.html} {Tree of thoughts: Deliberate problem solving with large language models}.
\newblock In \emph{Advances in Neural Information Processing Systems 36: Annual Conference on Neural Information Processing Systems 2023, NeurIPS 2023, New Orleans, LA, USA, December 10 - 16, 2023}.

\bibitem[{Yu et~al.(2024{\natexlab{a}})Yu, Zhang, Tiwari, and Wang}]{DBLP:journals/csur/YuZTW24}
Fei Yu, Hongbo Zhang, Prayag Tiwari, and Benyou Wang. 2024{\natexlab{a}}.
\newblock \href {https://doi.org/10.1145/3664194} {Natural language reasoning, {A} survey}.
\newblock \emph{{ACM} Comput. Surv.}, 56(12):304:1--304:39.

\bibitem[{Yu et~al.(2024{\natexlab{b}})Yu, Zhang, and Feng}]{DBLP:journals/corr/abs-2411-19443}
Tian Yu, Shaolei Zhang, and Yang Feng. 2024{\natexlab{b}}.
\newblock \href {https://doi.org/10.48550/ARXIV.2411.19443} {Auto-rag: Autonomous retrieval-augmented generation for large language models}.
\newblock \emph{CoRR}, abs/2411.19443.

\bibitem[{Yuan et~al.(2024)Yuan, Yang, Wang, Zhao, and Liu}]{DBLP:conf/emnlp/YuanYWZL24}
Xiaowei Yuan, Zhao Yang, Yequan Wang, Jun Zhao, and Kang Liu. 2024.
\newblock \href {https://aclanthology.org/2024.emnlp-main.997} {Improving zero-shot {LLM} re-ranker with risk minimization}.
\newblock In \emph{Proceedings of the 2024 Conference on Empirical Methods in Natural Language Processing, {EMNLP} 2024, Miami, FL, USA, November 12-16, 2024}, pages 17967--17983. Association for Computational Linguistics.

\bibitem[{Zhai and Lafferty(2001)}]{DBLP:conf/sigir/ZhaiL01}
ChengXiang Zhai and John~D. Lafferty. 2001.
\newblock \href {https://doi.org/10.1145/383952.384019} {A study of smoothing methods for language models applied to ad hoc information retrieval}.
\newblock In \emph{{SIGIR} 2001: Proceedings of the 24th Annual International {ACM} {SIGIR} Conference on Research and Development in Information Retrieval, September 9-13, 2001, New Orleans, Louisiana, {USA}}, pages 334--342. {ACM}.

\bibitem[{Zhai and Lafferty(2006)}]{DBLP:journals/ipm/ZhaiL06}
ChengXiang Zhai and John~D. Lafferty. 2006.
\newblock \href {https://doi.org/10.1016/J.IPM.2004.11.003} {A risk minimization framework for information retrieval}.
\newblock \emph{Inf. Process. Manag.}, 42(1):31--55.

\bibitem[{Zhang et~al.(2024{\natexlab{a}})Zhang, Zhoubian, Hu, Yue, Dong, and Tang}]{DBLP:conf/nips/ZhangZHYD024}
Dan Zhang, Sining Zhoubian, Ziniu Hu, Yisong Yue, Yuxiao Dong, and Jie Tang. 2024{\natexlab{a}}.
\newblock \href {http://papers.nips.cc/paper\_files/paper/2024/hash/76ec4dc30e9faaf0e4b6093eaa377218-Abstract-Conference.html} {Rest-mcts*: {LLM} self-training via process reward guided tree search}.
\newblock In \emph{Advances in Neural Information Processing Systems 38: Annual Conference on Neural Information Processing Systems 2024, NeurIPS 2024, Vancouver, BC, Canada, December 10 - 15, 2024}.

\bibitem[{Zhang et~al.(2024{\natexlab{b}})Zhang, Wang, Ren, Jiang, Wang, and Liu}]{DBLP:journals/corr/abs-2406-02746}
Jinghan Zhang, Xiting Wang, Weijieying Ren, Lu~Jiang, Dongjie Wang, and Kunpeng Liu. 2024{\natexlab{b}}.
\newblock \href {https://doi.org/10.48550/ARXIV.2406.02746} {{RATT:} {A} thought structure for coherent and correct {LLM} reasoning}.
\newblock \emph{CoRR}, abs/2406.02746.

\bibitem[{Zhao et~al.(2023{\natexlab{a}})Zhao, Li, Joty, Qin, and Bing}]{DBLP:conf/acl/ZhaoLJQB23}
Ruochen Zhao, Xingxuan Li, Shafiq Joty, Chengwei Qin, and Lidong Bing. 2023{\natexlab{a}}.
\newblock \href {https://doi.org/10.18653/V1/2023.ACL-LONG.320} {Verify-and-edit: {A} knowledge-enhanced chain-of-thought framework}.
\newblock In \emph{Proceedings of the 61st Annual Meeting of the Association for Computational Linguistics (Volume 1: Long Papers), {ACL} 2023, Toronto, Canada, July 9-14, 2023}, pages 5823--5840. Association for Computational Linguistics.

\bibitem[{Zhao et~al.(2023{\natexlab{b}})Zhao, Zhou, Li, Tang, Wang, Hou, Min, Zhang, Zhang, Dong, Du, Yang, Chen, Chen, Jiang, Ren, Li, Tang, Liu, Liu, Nie, and Wen}]{DBLP:journals/corr/abs-2303-18223}
Wayne~Xin Zhao, Kun Zhou, Junyi Li, Tianyi Tang, Xiaolei Wang, Yupeng Hou, Yingqian Min, Beichen Zhang, Junjie Zhang, Zican Dong, Yifan Du, Chen Yang, Yushuo Chen, Zhipeng Chen, Jinhao Jiang, Ruiyang Ren, Yifan Li, Xinyu Tang, Zikang Liu, and 3 others. 2023{\natexlab{b}}.
\newblock \href {https://doi.org/10.48550/ARXIV.2303.18223} {A survey of large language models}.
\newblock \emph{CoRR}, abs/2303.18223.

\end{thebibliography}
\clearpage
\appendix
\label{sec:appendix}

\section{Impact of Retrieval System Quality}

To evaluate how sensitive \sys is to the quality of the underlying retrieval system, we examine two key factors: the type of retriever and the number of retrieved documents.

\paragraph{Choice of Retriever.}
\begin{table}[h]
\centering
\small
\renewcommand{\arraystretch}{1.2}
\setlength{\tabcolsep}{3pt}
\begin{tabular}{lcccc}
\toprule
\textbf{Dataset} & \textbf{BM25 EM} & \textbf{BM25 F1} & \textbf{BGE EM} & \textbf{BGE F1} \\
\midrule
Musique   & 40.50 & 65.87 & 44.50 & 70.47 \\
HotpotQA  & 73.50 & 75.39 & 74.00 & 75.41 \\
2Wiki     & 56.50 & 62.61 & 72.00 & 67.36 \\
\bottomrule
\end{tabular}
\caption{\textbf{Impact of Retriever Type.} The dense retriever (BGE) outperforms the sparse retriever (BM25) across all datasets, with more notable gains on challenging benchmarks.}
\label{tab:retriever_sensitivity}
\end{table}

We compare the performance of \sys when paired with a sparse retriever (BM25) versus a dense retriever (BGE). 
As summarized in \autoref{tab:retriever_sensitivity}, dense retrieval consistently outperforms sparse retrieval across all three datasets. 
On average, BGE improves EM and F1 scores by 11.7\%, with particularly substantial gains observed on more challenging datasets such as 2Wiki. 
This indicates that while \sys exhibits some robustness to retrieval quality on simpler tasks (\eg HotpotQA), it substantially benefits from higher-quality evidence in more complex scenarios. 
These results suggest that although the verifier-guided reasoning in \sys can partially mitigate retrieval noise, it still relies on access to relevant and precise information to perform effective multi-hop reasoning.

\paragraph{Number of Retrieved Documents.}
\begin{table}[h]
\centering
\small
\renewcommand{\arraystretch}{1.2}
\setlength{\tabcolsep}{6pt}
\begin{tabular}{c c c}
\toprule
\textbf{\#Doc} & \textbf{EM} & \textbf{F1} \\
\midrule
1 & 28.50 & 44.14 \\
2 & 40.50 & 65.87 \\
3 & 37.00 & 63.97 \\
\bottomrule
\end{tabular}
\caption{\textbf{Effect of the Number of Retrieved Documents.} Performance peaks at retrieving two documents, suggesting a balance between sufficient evidence and retrieval noise.}
\label{tab:numdoc_sensitivity}
\end{table}

We further assess the impact of varying the number of retrieved documents on the Musique dataset. 
As shown in \autoref{tab:numdoc_sensitivity}, retrieving two documents yields the best performance. 
Adding a third document does not lead to further improvements and may even slightly degrade performance. 
This can be attributed to two factors. 
First, the knowledge density of the task affects how much additional context is beneficial.
Second, due to \sys's step-by-step reasoning design, the overall problem is decomposed into smaller, more manageable subproblems, each requiring less contextual information. 
In a nutshell, \sys is more sensitive to missing critical information than to receiving redundant or noisy evidence, exhibiting a degree of robustness to retrieval noise.

\section{Further Details}

\subsection{Further Details for Datasets}
\label{sec:appd}
We use HotpotQA~\citep{DBLP:conf/emnlp/Yang0ZBCSM18}, 2WikiMultihopQA~\citep{DBLP:conf/coling/HoNSA20}, and MusiQue~\citep{DBLP:journals/tacl/TrivediBKS22} as the test set, 
which are three representative benchmarks for open-ended, knowledge-intensive, and complex reasoning tasks.
The questions in these datasets require retrieving and reasoning over multiple supporting documents to answer, and they cover a wide range of topics without being constrained by any existing knowledge base or schema.
We performed preprocessing on the dataset. 
Specifically, considering the limitations of computational resources, we randomly sampled 200 questions from each dataset as the final test set. 
Each instance includes the original question and its answer, along with possible reference documents and necessary supporting documents. 
During the testing phase, we treated the possible reference documents for each question as its external knowledge base and employed BM25 as our retriever, with each retrieval returning the top two documents. 
Based on the original dataset, the majority of questions involve complex reasoning with three or more hops (over 80\%)~\citep{feng2025airrag}. 
Among them, Musique has the highest reasoning difficulty, with a notable higher proportion of multi-hop questions. 
We fixed the search depth to four layers to align with the number of sub-question decompositions required for problem-solving, thereby reducing unnecessary reasoning overhead. 
Additionally, we set the initial maximum number of expandable child nodes to 5 to cover hop counts comprehensively, guaranteeing diversity in the model's decomposition of questions.
As the search depth increases, the diversity of decomposition perspectives for original questions gradually decreases. 
Therefore, we also progressively reduced the number of expandable child nodes, which ensures search efficiency.

\subsection{Further Details for Evaluations}
\label{sec:appm}
We choose EM accuracy and F1 score as the evaluation metrics. 
EM accuracy measures the success rate of the results, while F1 score evaluates the reliability of the reasoning process. 
Specifically, for EM, we adopt the covered EM calculation method. 
We preprocess both the model's predictions and the ground truth answers by converting them to a uniform case format.
If the ground truth answer is a substring of the predicted result, we consider the prediction correct. 
This approach aims to genuinely reflect the method's performance. 
Additionally, we employ concise response prompts to ensure the model's final output is not overly verbose, thereby avoiding false-positive cases.
For the calculation of F1 score, we utilize the top two documents related to the entire reasoning path and original question. 
At the final step, we perform an additional retrieval based on the complete reasoning state to ensure the documents reflect the context of the entire path.
To prevent redundant retrievals from obscuring the comparison of F1 scores, we limit each retrieval to return only the top two documents.

\subsection{Further Details for LLMs}
Throughout the experiments, we primarily utilized the Qwen series~\citep{qwen2.5} and Llama series models~\citep{llama}. 
The Qwen series includes models with scales of 0.5B, 3B, 7B, 14B, and 32B parameters, while the experiments with Llama were mainly conducted on the 8B parameter model. 
To ensure fairness in the experiments, for any tasks involving risk assessment or state evaluation, we consistently employed the corresponding policy models.
In addition, experiments also involved DeepSeek-R1 distilled models~\citep{deepseek-r1}, including Qwen-7B, Qwen-14B, Qwen-32B, and Llama-8B. 
The distilled small model extracts reasoning patterns from DeepSeek-R1 and demonstrates superior performance compared to the inference patterns obtained through reinforcement learning~\citep{deepseek-r1}.

\subsection{Further Details for Prompts}
\label{sec:appp}
We list the full details of all prompts employed in \sys as follows.
The prompts used for forward inference follow a zero-shot instruction to directly describe the task.
We provide few-shot demonstrations in the prompts for risk assessment.

\clearpage
\onecolumn
\begin{tcolorbox}[
  width=1\textwidth,
  colback=gray!5,
  colframe=black,
  title=\texttt{Problem Decomposition},
  coltitle=black,
  colbacktitle=lightfreshblue,
  fonttitle=\large\bfseries,
  boxrule=0.8pt,
  arc=4pt,
  left=4pt,
  right=4pt,
  top=4pt,
  bottom=4pt,
]

Your task is to decompose the original question into one smaller sub-question based on the Intermediate answer and Observation.

The decomposed process is encouraged to be done from multiple perspectives.

Output a thought to reason the original question, and output one sub-question that you think is appropriate to solve next.

\textbf{DO NOT REPEAT} the question and \textbf{DO NOT} try to answer the question.

The output format is limited to:\\
Thought: ...\\
Sub-question: ...\\
Here, the "..." indicates omitted output information that you need to fill in.

\vspace{1em}
\textbf{Original question:} \textcolor{orange}{\{original question\}}

\textbf{Intermediate answers:} \textcolor{orange}{\{reasoning state\}}

\textbf{Observation:} \textcolor{orange}{\{retrieved documents\}}

\textbf{Output:}
\end{tcolorbox}
\begin{tcolorbox}[
  width=1\textwidth,
  colback=gray!5,
  colframe=black,
  title=\texttt{Intermediate Answer Generation},
  coltitle=black,
  colbacktitle=lightfreshblue,
  fonttitle=\large\bfseries,
  boxrule=0.8pt,
  arc=4pt,
  left=4pt,
  right=4pt,
  top=4pt,
  bottom=4pt,
]

Your task is to answer the following question using provided supporting facts.

The output answer should be a complete declarative sentence, rather than directly outputting phrases or words.

\textbf{DO NOT} use pronouns in the sentence.

Specially, if no provided supporting facts, just output "No directly relevant facts found." and nothing else.

\vspace{1em}
\textbf{Question:} \textcolor{orange}{\{sub-question\}}

\textbf{Supporting facts:} \textcolor{orange}{\{retrieved documents\}}

\textbf{Output:}
\end{tcolorbox}
\begin{tcolorbox}[
  width=1\textwidth,
  colback=gray!5,
  colframe=black,
  title=\texttt{Final Answer Generation},
  coltitle=black,
  colbacktitle=lightfreshblue,
  fonttitle=\large\bfseries,
  boxrule=0.8pt,
  arc=4pt,
  left=4pt,
  right=4pt,
  top=4pt,
  bottom=4pt,
]

Your task is to answer the original question based on the intermediate answers.

Output the final answer directly and nothing else.

\vspace{1em}
\textbf{Original question:} \textcolor{orange}{\{original question\}}

\textbf{Intermediate answers:} \textcolor{orange}{\{reasoning state\}}

\textbf{Output:}
\end{tcolorbox}
\begin{tcolorbox}[
  width=1\textwidth,
  colback=gray!5,
  colframe=black,
  title=\texttt{Risk Assessment},
  coltitle=black,
  colbacktitle=lightfreshblue,
  fonttitle=\large\bfseries,
  boxrule=0.8pt,
  arc=4pt,
  left=4pt,
  right=4pt,
  top=4pt,
  bottom=4pt,
]

Given intermediate answer containing the facts about the original question, which is unknown, your task is to infer what the original question might have been.

Output the most likely original question directly and nothing else.

\textbf{Example 1:}

Intermediate answer:

Muhammad Ali was 74 years old when he died.

Alan Turing was 41 years old when he died.

The original question might be:

Who lived longer, Muhammad Ali or Alan Turing?

\textbf{Example 2:}

Intermediate answer:

Craigslist was founded by Craig Newmark. 

Craig Newmark was born on December 6, 1952. 

The original question might be:

When was the founder of craigslist born?

\vspace{1em}
\textbf{Intermediate answers:} \textcolor{orange}{\{reasoning state\}}

\textbf{The original question might be:}
\end{tcolorbox}
\begin{tcolorbox}[
  width=1\textwidth,
  colback=gray!5,
  colframe=black,
  title=\texttt{LLM-as-Verifier},
  coltitle=black,
  colbacktitle=lightfreshblue,
  fonttitle=\large\bfseries,
  boxrule=0.8pt,
  arc=4pt,
  left=4pt,
  right=4pt,
  top=4pt,
  bottom=4pt,
]

Given a question, your task is to determine the consistency score of its decomposed sub-questions and corresponding intermediate answers with the original question.

Directly output \textbf{JUST A NUMBER} between 0 and 10 to represent the consistency score.

\textbf{DO NOT} output anything else.

\vspace{1em}
\textbf{Original question:} \textcolor{orange}{\{original question\}}

\textbf{Sub-questions:} \textcolor{orange}{\{sub-questions\}}

\textbf{Intermediate answers:} \textcolor{orange}{\{reasoning state\}}

\textbf{Output:}
\end{tcolorbox}
\begin{tcolorbox}[
  width=1\textwidth,
  colback=gray!5,
  colframe=black,
  title=\sys \texttt{Simple Case},
  coltitle=black,
  colbacktitle=lightfreshblue,
  fonttitle=\large\bfseries,
  boxrule=0.8pt,
  arc=4pt,
  left=4pt,
  right=4pt,
  top=4pt,
  bottom=4pt,
]
\textbf{STATE 1:} 
depth: 1; action: decompose\\
original question: When was the last time Peter Till's team beat winner of 1894-95 FA Cup in SC?\\
sub-questions: Who was the winner of FA Cup?\\
intermediate-answer: \\
reasoning state: 

\textbf{STATE 2:} 
depth: 1; action: retrieve-then-reason\\
original question: When was the last time Peter Till's team beat winner of 1894-95 FA Cup in SC?\\
sub-questions: Who was the winner of FA Cup?\\
-- retrieve: \textcolor{orange}{\{supporting facts\}}\\
intermediate-answer: The winner of the 1894- 95 FA Cup is Aston Villa.\\
reasoning state: The winner of the 1894- 95 FA Cup is Aston Villa.

\textbf{STATE 3:} 
depth: 2; action: decompose\\
original question: When was the last time Peter Till's team beat winner of 1894-95 FA Cup in SC?\\
sub-questions: What was the sports team of Peter Till?\\
intermediate-answer: \\
reasoning state: The winner of the 1894- 95 FA Cup is Aston Villa.

\textbf{STATE 4:} 
depth: 2; action: retrieve-then-reason\\
original question: When was the last time Peter Till's team beat winner of 1894-95 FA Cup in SC?\\
sub-questions: What was the sports team of Peter Till?\\
-- retrieve: \textcolor{orange}{\{supporting facts\}}\\
intermediate-answer: Peter Till’s sports team is Birming-ham City.\\
reasoning state: The winner of the 1894- 95 FA Cup is Aston Villa. Peter Till’s sports team is Birming-ham City.

\textbf{STATE 5:} 
depth: 3; action: decompose\\
original question: When was the last time Peter Till's team beat winner of 1894-95 FA Cup in SC?\\
sub-questions: When was the last time Birming-ham City beat Aston Villa in SC?\\
intermediate-answer: \\
reasoning state: The winner of the 1894- 95 FA Cup is Aston Villa. Peter Till’s sports team is Birming-ham City.

\textbf{STATE 6:} 
depth: 3; action: retrieve-then-reason\\
original question: When was the last time Peter Till's team beat winner of 1894-95 FA Cup in SC?\\
sub-questions: When was the last time Birming-ham City beat Aston Villa in SC?\\
-- retrieve: \textcolor{orange}{\{supporting facts\}}\\
intermediate-answer: The last time Birming-ham City beat Aston Villa was in March 2005.\\
reasoning state: The winner of the 1894- 95 FA Cup is Aston Villa. Peter Till’s sports team is Birming-ham City. The last time Birming-ham City beat Aston Villa was in March 2005. 

\textbf{STATE 7:} 
depth: 4; action: answer\\
original question: When was the last time Peter Till's team beat winner of 1894-95 FA Cup in SC?\\
sub-questions: \\
intermediate-answer: \\
reasoning state: The winner of the 1894- 95 FA Cup is Aston Villa. Peter Till’s sports team is Birming-ham City. The last time Birming-ham City beat Aston Villa was in March 2005. \\
\textcolor{red}{final answer}: March 2005

\vspace{1em}
\textbf{NOTE:} Decomposition is based on original question, reasoning state, and retrieved documents.

\end{tcolorbox}

\end{document}